\pdfoutput=1

\documentclass[11pt]{article}

\usepackage[final]{acl}

\usepackage{times}
\usepackage{latexsym}
\usepackage{authblk}
\usepackage{adjustbox}

\usepackage[T1]{fontenc}

\usepackage[utf8]{inputenc}

\usepackage{microtype}

\usepackage{inconsolata}

\usepackage{graphicx}
\usepackage{amsmath}
\usepackage{multirow}
\usepackage{arydshln}
\usepackage{booktabs}
\usepackage{tcolorbox}
\usepackage{lineno}
\usepackage{bm}

\definecolor{darkred}{RGB}{139,0,0}
\definecolor{darkgreen}{RGB}{0,100,0}

\title{CoTKR: Chain-of-Thought Enhanced Knowledge Rewriting for Complex Knowledge Graph Question Answering}

\author{
 \vspace{-0.3cm}
 \textbf{Yike Wu\textsuperscript{1,2,${\ast}$,\dag} \thanks{${\ast}$ Equal contribution. \dag Corresponding author. }},
  \textbf{Yi Huang\textsuperscript{3,${\ast}$,\dag}},
  \textbf{Nan Hu\textsuperscript{1,2,${\ast}$,\dag}},\\
  
  \textbf{Yuncheng Hua\textsuperscript{4}},
 \textbf{Guilin Qi\textsuperscript{1,2}},
 \textbf{Jiaoyan Chen\textsuperscript{5}},
 \textbf{Jeff Z. Pan\textsuperscript{6\dag}}\\
  \textsuperscript{1}Southeast University, Nanjing, Jiangsu, China\\
  \textsuperscript{2}Key Laboratory of New Generation Artificial Intelligence Technology and\\Its Interdisciplinary Applications (Southeast University), Ministry of Education\\
  \textsuperscript{3}China Mobile Research Institute, Beijing, China 
  \textsuperscript{4}Monash University, Melbourne, Australia\\
  \textsuperscript{5}University of Manchester, Manchester, UK 
  \textsuperscript{6}University of Edinburgh, Edinburgh, UK\\
  \texttt{\{yike.wu,nanhu\}@seu.edu.cn}, \texttt{huangyi@chinamobile.com} 
  }

\makeatletter
\def\thanks#1{\protected@xdef\@thanks{\@thanks
\protect\footnotetext{#1}}}
\makeatother

\begin{document}
\maketitle

\begin{abstract}
Recent studies have explored the use of Large Language Models (LLMs) with Retrieval Augmented Generation (RAG) for Knowledge Graph Question Answering (KGQA). They typically require rewriting retrieved subgraphs into natural language formats comprehensible to LLMs.
However, when tackling complex questions, the knowledge rewritten by existing methods may include irrelevant information, omit crucial details, or fail to align with the question's semantics. To address them, we propose a novel rewriting method CoTKR, \textbf{C}hain-\textbf{o}f-\textbf{T}hought Enhanced \textbf{K}nowledge \textbf{R}ewriting, for generating reasoning traces and corresponding knowledge in an interleaved manner, thereby mitigating the limitations of single-step knowledge rewriting. Additionally, to bridge the preference gap between the knowledge rewriter and the question answering (QA) model, we propose a training strategy PAQAF, \textbf{P}reference \textbf{A}lignment from \textbf{Q}uestion \textbf{A}nswering \textbf{F}eedback, 
for leveraging feedback from the QA model to further optimize the knowledge rewriter. We conduct experiments using various LLMs across several KGQA benchmarks. Experimental results demonstrate that, compared with previous knowledge rewriting methods, CoTKR generates the most beneficial knowledge representation for QA models, which significantly improves the performance of LLMs in KGQA \footnote{Our code is available at https://github.com/wuyike2000/CoTKR.}.
\end{abstract}

\section{Introduction}

Large Language Models (LLMs) have achieved remarkable performance across various natural language processing tasks, marking a significant milestone \cite{DBLP:conf/iclr/SanhWRBSACSRDBX22,DBLP:conf/nips/BrownMRSKDNSSAA20,DBLP:journals/corr/abs-2205-01068,DBLP:journals/dint/AzariaAR24,publisher/Beijing}. Despite their superior performance in zero-shot scenarios \cite{DBLP:conf/iclr/WeiBZGYLDDL22,DBLP:conf/nips/KojimaGRMI22}, they still encounter factual errors, known as ``hallucinations'' \cite{DBLP:journals/csur/JiLFYSXIBMF23}, especially in knowledge-intensive tasks like question answering (QA) \cite{DBLP:journals/www/HuWQMCPA23,DBLP:journals/corr/abs-2303-07992,DBLP:journals/dint/LiZLYC23,HUGP2023}. This issue arises due to the intrinsic limitations of LLMs, including factual inaccuracies and outdated knowledge~\cite{PRKSC2023}. To address this challenge, a substantial of work \cite{DBLP:conf/emnlp/MaGHZD23,DBLP:conf/acl/TrivediBKS23,DBLP:journals/corr/abs-2309-11206} retrieves task-relevant knowledge from external sources as context, thereby enhancing the capabilities of LLMs in downstream tasks, known as Retrieval-Augmented Generation (RAG) \cite{DBLP:conf/nips/LewisPPPKGKLYR020,DBLP:journals/corr/abs-2312-10997,HLVPP2023}.

\begin{figure*}[h]
\centering
  \includegraphics[width=0.9\linewidth]{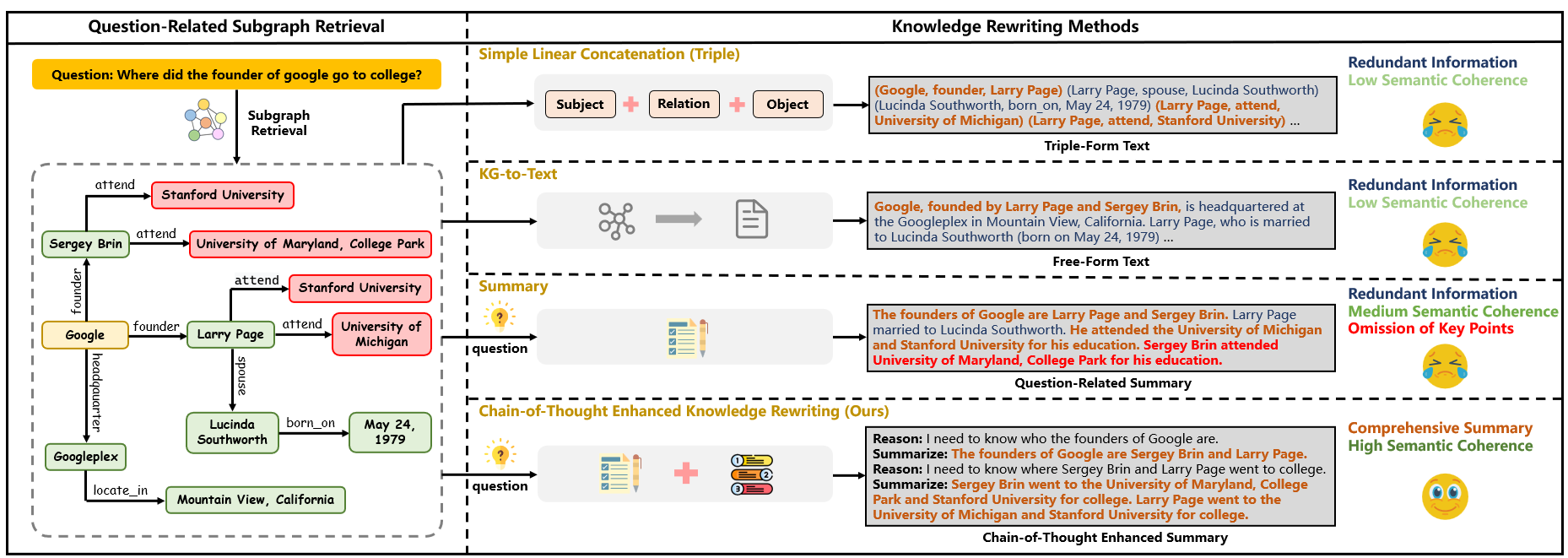}
  \caption{The commonly used knowledge rewriting methods in existing work.}
  \label{fig 1}
\end{figure*}

Recent work \cite{DBLP:journals/corr/abs-2309-11206,baek-etal-2023-knowledge,sen-etal-2023-knowledge,DBLP:journals/corr/abs-2401-00426} under the RAG paradigm explores the use of Knowledge Graphs (KGs)~\cite{PVGW2017,PCEH+2017} as an information source to enhance the capabilities of LLMs in Question Answering (QA). Unlike typical QA tasks, a key challenge in KGQA under this paradigm lies in transforming question-related subgraphs into natural language that LLMs can understand while preserving the structural information \cite{DBLP:journals/corr/abs-2403-02966,DBLP:conf/www/DingLLQ24,DBLP:journals/corr/abs-2309-11206}. This process is referred to as Knowledge Rewriting (KR) in this study. As illustrated in Figure \ref{fig 1}, this paper summarizes the commonly used knowledge rewriting methods in existing work. Most prior studies \cite{baek-etal-2023-knowledge,sen-etal-2023-knowledge,DBLP:journals/corr/abs-2308-13259} employ simple linear concatenation method (Triple), which concatenates the subject, relation, and object of each triple to form triple-form text. Additionally, considering that LLMs are pre-trained on text corpora and struggle with structured triple-form text, some efforts \cite{DBLP:journals/corr/abs-2309-11206,DBLP:conf/aaai/BianH0021,DBLP:conf/jist/0007HCGFP0Z22} focus on converting triples into natural language through KG-to-Text. Furthermore, given that retrieved subgraphs often contain redundant information irrelevant to the question, other studies \cite{DBLP:journals/corr/abs-2403-02966,DBLP:conf/aaaiss/DernbachAZHC24} aim to extract question-relevant knowledge from the triples to generate summary pertinent to the question.

Although these strategies are effective, they exhibit several limitations: \textbf{(1) Redundancy or omissions.} As illustrated in Figure \ref{fig 1}, knowledge generated by \textsf{Triple} and \textsf{KG-to-Text} are verbose, containing excessive irrelevant information. \textsf{Summary} provides a question-related summary but attempts to organize all relevant knowledge in one step. Given the extensive knowledge necessary to address complex questions, this method may not encapsulate all critical information, potentially resulting in the omission of key points. \textbf{(2) Semantic mismatch.} The three existing methods shown in Figure \ref{fig 1} ignore the semantics of the question and lack a logical organization that aligns with the question's reasoning path.

To this end, we propose \textbf{C}hain-\textbf{o}f-\textbf{T}hought Enhanced \textbf{K}nowledge \textbf{R}ewriting, CoTKR. Inspired by ReAct \cite{DBLP:conf/iclr/YaoZYDSN023}, the core of our method involves generating reasoning traces and corresponding knowledge in an interleaved manner. As shown in Figure \ref{fig 1}, we alternate the following two operations: (1) \textbf{Reasoning}: decomposing the question to identify the knowledge required for inference; (2) \textbf{Summarization}: summarizing the relevant knowledge from retrieved triples, informed by the reasoning step's output. By integrating Chain-of-Thought (CoT) \cite{DBLP:conf/nips/Wei0SBIXCLZ22} with knowledge rewriting, CoTKR filters out irrelevant information and extracts question-related knowledge. Moreover, it generates a well-organized knowledge representation\footnote{In this paper, ``knowledge representation'' refers to the natural language form of question-related knowledge.} semantically aligned with the question. Unlike traditional CoT applications in QA, our framework employs the knowledge rewriter to first summarize knowledge, which then serves as contextual information to enhance QA performance. This strategy offers superior robustness. Although the summary might be inaccurate, it still contributes valuable information, potentially leading to correct answers. However, applying CoT to QA requires more precise reasoning chains, which are significantly affected by the error propagation \cite{DBLP:journals/corr/abs-2308-13259,DBLP:conf/iclr/YaoZYDSN023}. To train knowledge rewriters based on LLMs, we design a training framework for CoTKR. In the first stage, inspired by previous work \cite{DBLP:conf/emnlp/MaGHZD23,DBLP:journals/corr/abs-2309-11206,DBLP:journals/corr/abs-2403-02966}, we use knowledge representations generated by ChatGPT to guide the supervised fine-tuning of the knowledge rewriter, enabling it to initially master the capability of knowledge rewriting. In the second stage, we introduce \textbf{P}reference \textbf{A}lignment from \textbf{Q}uestion \textbf{A}nswering \textbf{F}eedback (PAQAF) to bridge the preference gap between the knowledge rewriter and the QA model. This method evaluates the quality of different knowledge representations based on the corresponding responses from the QA model. Subsequently, it constructs preference pairs, and fine-tunes LLMs through direct preference optimization (DPO) \cite{DBLP:conf/nips/RafailovSMMEF23}.

We conduct experiments on GrailQA \cite{DBLP:conf/www/GuKVSLY021} and GraphQuestions \cite{DBLP:conf/emnlp/SuSSSGYY16}, comparing commonly used knowledge rewriting methods in existing work. Contrary to previous findings \cite{DBLP:journals/corr/abs-2402-11541,baek-etal-2023-knowledge}, which suggest that LLMs perform better with knowledge in triple-form rather than in natural language, our findings demonstrate that LLMs can significantly benefit from knowledge represented in carefully crafted natural language. This indicates that our method could substantially enhance the performance of LLMs in KGQA.

The main contributions of this paper are:

\begin{itemize}

\item We propose CoTKR, a \textbf{C}hain-\textbf{o}f-\textbf{T}hought Enhanced \textbf{K}nowledge \textbf{R}ewriting method to improve the quality of knowledge representation through the application of CoT. This method generates reasoning traces and corresponding knowledge in an interleaved manner, thereby producing well-organized knowledge representations that are coherent with the question's semantics.

\item We propose a training strategy PAQAF, \textbf{P}reference \textbf{A}lignment from \textbf{Q}uestion \textbf{A}nswering \textbf{F}eedback, to bridge the preference gap between the knowledge rewriter and the QA model. It assesses the quality of different knowledge representations by evaluating corresponding responses from the QA model. Then, it constructs preference pairs and employs DPO to optimize the knowledge rewriter.

\item We conduct experiments on two KGQA benchmarks. Compared with other knowledge rewriting methods, CoTKR can generate the most beneficial knowledge representation for the QA model and further enhance the performance of LLMs in KGQA. Additionally, considering privacy and cost issues, we evaluate the performance of open-source and closed-source LLMs as the foundational models for knowledge rewriting and QA.

\end{itemize}

\section{Related Work}

\subsection{KG-Augmented LLMs for KGQA}

To mitigate hallucination in LLMs, existing work \cite{DBLP:journals/corr/abs-2309-11206,baek-etal-2023-knowledge,sen-etal-2023-knowledge,DBLP:journals/corr/abs-2401-00426} attempts to enhance LLMs with KGs in the RAG paradigm. The na\"{i}ve approach involves retrieving question-related triples from KGs as contextual information for QA \cite{baek-etal-2023-knowledge,sen-etal-2023-knowledge}. Although this method has proven effective, there remains ample room for improvement. Some studies \cite{DBLP:journals/corr/abs-2401-00426,DBLP:journals/corr/abs-2308-13259} integrate Chains-of-Thought (CoT) \cite{DBLP:conf/nips/Wei0SBIXCLZ22} with RAG to tackle complex questions. Keqing\cite{DBLP:journals/corr/abs-2401-00426} decomposes complex questions using predefined templates, retrieves candidate entities from KG, reasons through sub-questions, and ultimately generates answers with clear reasoning paths. KD-CoT \cite{DBLP:journals/corr/abs-2308-13259} validates and adjusts reasoning traces in CoT through interactions with external knowledge, thereby addressing issues of hallucinations and error propagation. Furthermore, alternative efforts \cite{DBLP:journals/corr/abs-2309-11206,DBLP:journals/corr/abs-2403-02966} address the limitations of LLMs in processing structured knowledge and the noise in retrieved triples by post-processing these triples into natural language or summaries pertinent to the questions.

This paper focuses on optimizing the knowledge representation under the RAG paradigm for KGQA. Unlike previous work that transforms triples into the natural language in one step, we adopt CoT to summarize relevant knowledge step-by-step, ensuring comprehensiveness and semantic coherence in the generated knowledge.

\subsection{Preference Alignment for LLMs on Question Answering}

LLMs have the potential to generate content that contains gender discrimination, unethical elements, and racial biases, inconsistent with human values \cite{DBLP:journals/corr/abs-2307-14192,ray2023chatgpt}. To address this issue, Preference Alignment (PA) \cite{DBLP:journals/corr/abs-2310-19852,DBLP:journals/corr/abs-2307-12966} aims to fine-tune LLMs to align with human preferences. Existing QA work based on LLMs uses PA to bridge the gap between model preferences and those of humans or the QA tasks. KnowPAT \cite{DBLP:journals/corr/abs-2311-06503} trains LLMs on a knowledge preference set to align their knowledge biases with human preferences, selecting better factual knowledge as context. BGM \cite{DBLP:journals/corr/abs-2401-06954} utilizes downstream task metrics as rewards to optimize the bridging model between retrievers and QA models. Rewrite-Retrieve-Read \cite{ma-etal-2023-query} employs QA evaluation metrics as reward signals, fine-tuning the query rewriting module. EFSUM \cite{DBLP:journals/corr/abs-2403-02966} constructs preference pairs sampled from LLMs and fine-tunes the knowledge summarizer using the Direct Preference Optimization (DPO) \cite{DBLP:conf/nips/RafailovSMMEF23} algorithm.

Our study innovatively employs responses from the QA model to evaluate the quality of knowledge representations. We then construct preference pairs from these evaluations and optimize the knowledge rewriter using DPO.

\begin{figure*}[h]
\centering
  \includegraphics[width=0.9\linewidth]{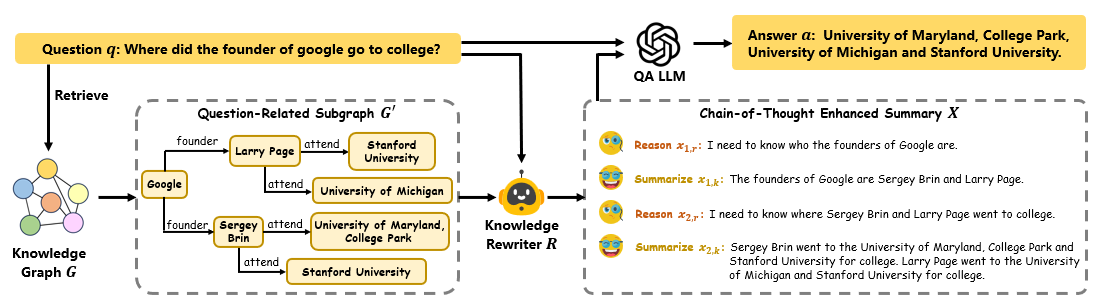}
  \caption{Illustration of our KGQA framework. CoTKR generates reasoning traces and corresponding knowledge in an interleaved manner.}
  \label{fig 2}
\end{figure*}

\section{Preliminaries}

\noindent\textbf{Knowledge Graph} (KG) is a structured collection of triples in the form of $(s,r,o)$, where $s$, $r$, $o$ represent the subject, the relation, and the object. This collection is denoted by $G=\{(s, r, o)\mid s, o \in E, r \in R\}$, where $E$ represents the set of entities and $R$ represents the set of relations.

\noindent\textbf{Knowledge Graph Question Answering} (KGQA) aims to answer natural language questions by utilizing a set of facts within KGs. Following previous work \cite{DBLP:conf/acl/SaxenaTT20,DBLP:conf/iclr/JiangZ0W23}, we assume that the subject entity of the question is given. Given a question $q$ and a subject entity $e$, the objective is to generate a response $a$ using the relevant facts in the KG $G$ that accurately addresses the question.

\noindent\textbf{Knowledge Rewriting} (KR) for KGQA aims to transform question-related triples into natural language that can be consumed by LLMs. Given a question $q$ and a subgraph $G'=\{(s,r,o)\mid s, o \in E, r \in R\}$ retrieved from KG $G$, the task is to generate a natural language sequence $X$ that provides contextual information to answer the question.

\section{Methods}

\subsection{Chain-of-Thought Enhanced Knowledge Rewriting}

The architecture of our QA framework is depicted in Figure \ref{fig 2}. Initially, our framework retrieves a question-related subgraph from the KG, which is subsequently transformed into contextual knowledge using CoTKR. This contextual knowledge, along with the question, prompts the QA model to generate an answer.  The core of this framework is the knowledge rewriter. 
Briefly, it alternatively conducts the following two operations: \textbf{Reasoning}: decomposing the question and generating a reasoning trace based on generated knowledge representation and pointing out the specific knowledge needed for the current step; \textbf{Summarization}: summarizing the relevant knowledge based on the current reasoning trace from the subgraph.



Assume we have the reasoning traces at step $t-1$ as $x_{t-1, r}$ and the summarized knowledge at step $t-1$ as $x_{t-1,k}$. The corresponding knowledge representation, i.e., $X_{t-1}$ is represented as:
\begin{equation}
X_{t-1}=[x_{1,r}, x_{1,k}, ..., x_{t-1,r}, x_{t-1,k}].
\end{equation}

For knowledge rewriting at step $t$, given the question $q$, the subgraph $G'$, and the previously generated content $X_{t-1}$, the knowledge rewriter $R$ first generates the reasoning trace $x_{t,r}$:
\begin{equation}
x_{t,r}=R(q,G',X_{t-1})
\end{equation}

Subsequently, based on the question $q$, the subgraph $G'$, the previously generated content $X_{t-1}$, and the reasoning trace at step $t$ $x_{t,r}$, CoTKR summarizes relevant knowledge $x_{t,k}$:
\begin{equation}
x_{t,k}=R(q,G',X_{t-1},x_{t,r})
\end{equation}

$x_{t,r}$ and $x_{t,k}$ are attached to $X_{t-1}$ for the knowledge representation at step $t$. Note in step $1$, $X_0$ is initialized to None.


\subsection{Training Framework for CoTKR}

\begin{figure}[h]
\centering
  \includegraphics[width=0.9\columnwidth]{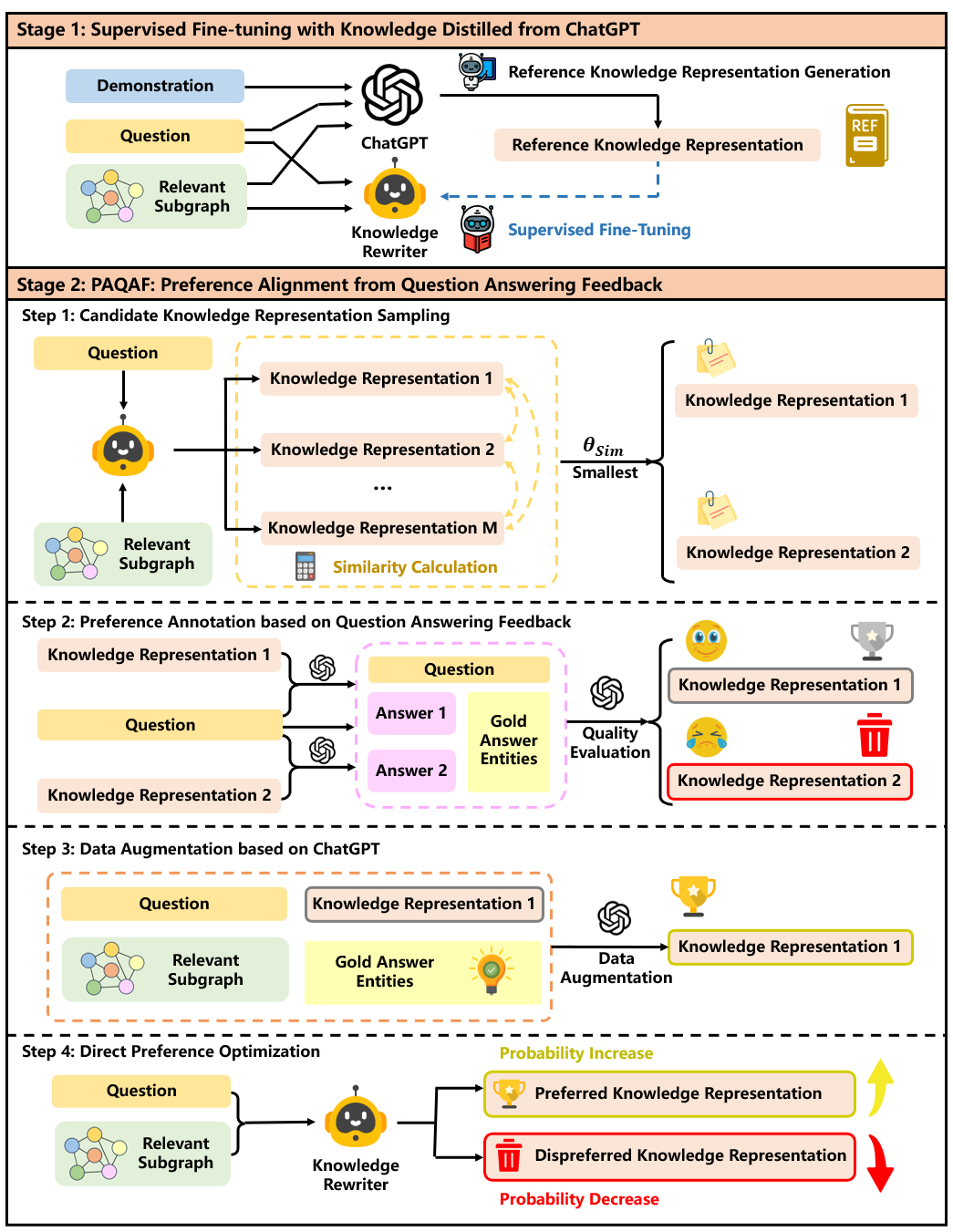}
  \caption{Our training framework for CoTKR.}
  \label{fig 3}
\end{figure}

Figure \ref{fig 3} illustrates the training framework of CoTKR.

\subsubsection{Supervised Fine-tuning with Knowledge Distilled from ChatGPT}

This stage enables open-source LLMs to initially acquire the knowledge rewriting capability through supervised fine-tuning. This primarily comprises two steps: reference knowledge representation generation and supervised fine-tuning.

\noindent\textbf{Reference Knowledge Representation Generation.} Inspired by previous work \cite{DBLP:conf/emnlp/MaGHZD23,DBLP:journals/corr/abs-2309-11206,DBLP:journals/corr/abs-2403-02966}, we employ ChatGPT as the data generator to construct training corpora. We verbalize the question-related subgraph, $G'$, through simple linear concatenation and combine it with the question, $q$, to form the input prompt $x$. Subsequently, ChatGPT generates the reference knowledge representation $k$ based on several examples (i.e., demonstrations) and the provided input $x$. Finally, we construct the training dataset $D_T=\{(x_1, k_1),(x_2,k_2), ..., (x_T,k_T)\}$.

\noindent\textbf{Supervised Fine-tuning.} For every pair of input and output $(x_i, k_i)$ in the training dataset $D_T$, our knowledge rewriter $R_{\theta}$ is trained to generate $k_i$ based on $x_i$ using the following objective:
\begin{equation}
L_{SFT}=-\frac{1}{T}\sum_{i=1}^{T}\log p_{\theta}(k_i|x_i)
\end{equation}
where $\theta$ represents the parameters of the knowledge rewriter $R_{\theta}$ and $p_{\theta}(k_i|x_i)$ signifies the probability that $R_{\theta}$ generates $k_i$, given the input $x_i$.

\subsubsection{Preference Alignment from Question Answering Feedback (PAQAF)}

In this stage, Preference Alignment (PA) is employed to bridge the preference gap between the knowledge rewriter and the QA model. This stage includes four steps: candidate knowledge representation sampling, preference annotation based on QA feedback, data augmentation based on ChatGPT, and direct preference optimization (DPO).

\noindent\textbf{Candidate Knowledge Representation Sampling.} We input the question $q$ and the corresponding subgraph $G'$, then sample $M$ candidate knowledge representations, $k_1, k_2, ..., k_M$, from the knowledge rewriter $R_{\theta}$.

\noindent\textbf{Preference Annotation based on Question Answering Feedback.} Among the candidate knowledge representations, we select the two, $k_1$ and $k_2$, with the greatest semantic difference (i.e., the lowest similarity) to facilitate faster convergence during training. Utilizing standard evaluation methods for assessing these representations is suboptimal, as they fail to align with the preferences of QA models. Inspired by the findings in previous work\cite{DBLP:journals/corr/abs-2309-11206,DBLP:journals/corr/abs-2403-02966,DBLP:journals/corr/abs-2311-06503}, we posit that better knowledge representations generally lead to better performance on QA. Consequently, we adopt $k_1$ and $k_2$ as contextual knowledge, prompting the QA model $Q$ to answer the question $q$, generating answers $a_1$ and $a_2$, respectively. Subsequently, we prompt ChatGPT to assess the quality of $a_1$ and $a_2$ from the perspectives of accuracy and relevance. This evaluation aims to identify the preferred knowledge representation $k^+$ and the dispreferred knowledge representation $k^-$. Details of the evaluation prompt are provided in Appendix \ref{PA}.

\noindent\textbf{Data Augmentation based on ChatGPT}. ChatGPT is able to produce higher-quality knowledge representations, compared with open-source LLMs. Therefore, in order to improve the quality of preferred knowledge representation and enhance the diversity of the training data, we employ ChatGPT to paraphrase $k^+$. In addition to the question $q$, the retrieved subgraph $G'$, and the preferred knowledge representation $k^+$, we also provide the answer entity $e$. This allows ChatGPT to organize relevant knowledge around $e$, ensuring that the rewritten knowledge covers key evidence. We concatenate the question $q$ and the textualized subgraph $G'$ using a prompt template as the input $x$, and use the paraphrased knowledge representation $k^{++}$ and $k^-$ as the preferred pair. Finally, we construct the preference dataset $P_N={(x_1, k_1^{++}, k_1^-), (x_2, k_2^{++}, k_2^-), ..., (x_N, k_N^{++}, k_N^-)}$. The prompt for knowledge augmentation is in Appendix \ref{data augmentation}.

\noindent\textbf{Direct Preference Optimization.}
We employ Direct Preference Optimization (DPO) on our knowledge rewriter, $R_{\theta}$, to develop a preference-tuned version, $R_{\theta^*}$. It minimizes the following objective:
\begin{equation}
\begin{split}
&L_{DPO}(\theta^*;\theta)=\\
&-\frac{1}{N}\sum_{i=1}^{N}\log\sigma[r(x_i, k_i^{++})-r(x_i, k_i^-)]
\end{split}
\end{equation}
\begin{equation}
r(x_i, k_i)=\frac{p_{\theta^*}(k_i|x_i)}{p_{\theta}(k_i|x_i)}
\end{equation}
Considering the varying preferences of different QA models, CoTKR is specifically trained for each QA model. Through the two stages of training, CoTKR tends to generate more favorable knowledge representation $k^{++}$ for each QA model, while avoiding unhelpful knowledge representation $k^-$.

\section{Experiments}

\subsection{Datasets}

\textbf{GrailQA} \cite{DBLP:conf/www/GuKVSLY021} is a challenging, large-scale multi-hop KGQA benchmark that consists of 64,331 questions (44,337 train, 6,763 dev, 13,231 test). The training and dev sets provide annotated SPARQL query and answer entities, while the test set comprises only the questions. For evaluation convenience, the dev set is used for testing.

\noindent\textbf{GraphQuestions} \cite{DBLP:conf/emnlp/SuSSSGYY16} is a characteristic-rich dataset for factoid question answering based on Freebase. It comprises 5,166 questions (2,771 train, 2,395 test). For each question, the dataset provides corresponding SPARQL query and answer entities.

\subsection{Large Language Models}

In this experiment, Llama-2 (7B) \cite{touvron2023llama2}, Llama-3 (8B) \cite{llama3modelcard}, and ChatGPT \footnote{https://api.openai.com/} are employed for knowledge rewriting, while ChatGPT and Mistral (7B) \cite{jiang2023mistral} are used for QA tasks. The details of these LLMs are provided in Appendix \ref{Large Language Models}.


\subsection{Baselines}
We compare \textbf{CoTKR} (without PAQAF) and \textbf{CoTKR+PAQAF} (\textbf{CoTKR+PA} for shortness) with other knowledge rewriting methods in KGQA:

\noindent\textbf{Simple linear concatenation (Triple)} \cite{baek-etal-2023-knowledge,sen-etal-2023-knowledge} concatenates the subject, predicate, and object of a triple to generate triple-form text. This method does not require additional models for knowledge rewriting.

\noindent\textbf{KG-to-Text} \cite{DBLP:journals/corr/abs-2309-11206} transforms facts into the free-form text for each relation path with a KG-to-Text model, addressing the limitations of LLMs in understanding structured triple-form text.

\noindent\textbf{Summary} \cite{DBLP:journals/corr/abs-2403-02966} converts triples into a question-relevant summary, alleviating the issue of redundant contextual knowledge.

To ensure a fair comparison, we employ the same corpus generation method for both the baselines and our method. All baselines undergo supervised fine-tuning without preference alignment.

\subsection{Retrieval Methods}
The retrieval module is not the focus of our research. Therefore, we adopt three commonly used retrieval methods. For detailed implementation, please refer to Appendix \ref{retrieve}.

\noindent\textbf{2-Hop.} We retain 30 triples from the 2-hop subgraph of the head entity, prioritizing those with higher semantic similarity to the question.

\noindent\textbf{BM25.} We follow the processing method in DecAF \cite{DBLP:conf/iclr/YuZNZL0HWWX23}, simply linearizing the 1-hop subgraph of the topic entity as the article. We take the top 30 triples corresponding to the candidate documents as the retrieval results.

\noindent\textbf{Ground Truth Subgraph (GS).} We modify the SPARQL queries from the datasets to obtain the ground truth subgraphs. These subgraphs represent the results of an ideal retriever.

\subsection{Evaluation Metrics}
Following previous work on generative KGQA \cite{DBLP:journals/corr/abs-2309-11206,baek-etal-2023-knowledge,DBLP:journals/corr/abs-2403-02966}, we adopt \textbf{Accuracy (Acc)} as one of our evaluation metrics. It measures whether the model's response includes at least one answer entity. For a dataset comprising $N$ questions, \textbf{Acc} is calculated as follows:
\begin{equation}
Acc=\frac{\sum_{i=1}^{N}Acc_{i}}{N}
\end{equation}
\begin{equation}
Acc_i =
\begin{cases} 
1, & \text{if at least one answer entity} \\
   & \text{appears in the response} \\
0, & \text{if no answer entity appears in} \\
   & \text{the response}
\end{cases}
\end{equation}

In order to comprehensively assess the performance of KGQA, we employ \textbf{Recall} to evaluate the proportion of correct answer entities present in the model’s response. For a dataset containing $N$ questions, \textbf{Recall} is calculated as follows: 
\begin{equation}
Recall=\frac{\sum_{i=1}^{N}Recall_{i}}{N}
\end{equation}
\begin{equation}
Recall_i =\frac{N_{appear}}{N_{total}}
\end{equation}
where $N_{appear}$ refers to the number of answer entities contained in the model's responses, and $N_{total}$ refers to the total number of the answer entities. Additionally, we consider utilizing \textbf{Exact Match (EM)} as an evaluation metric. Given that the responses generated by LLMs consist of multiple paragraphs, while the corresponding answers are entities, we adjust the traditional EM metric. Our modified EM metric assesses whether all answer entities are included in the model's responses. For a dataset consisting of $N$ questions, \textbf{EM} is calculated as follows: 
\begin{equation}
EM=\frac{\sum_{i=1}^{N}EM_{i}}{N}
\end{equation}
\begin{equation}
EM_i =
\begin{cases} 
1, & \text{if all answer entities} \\
   & \text{appear in the response} \\
0, & \text{other cases}
\end{cases}
\end{equation}


\subsection{Main Results}

\begin{table*}[h]
\centering
\small
\scalebox{0.8}{
\begin{tabular}{llccccccc}
    \toprule
     \multirow{2}{*}{\textbf{KR LLMs}} & \multirow{2}{*}{\textbf{Methods}} & \multicolumn{3}{c}{\textbf{GrailQA}} & &\multicolumn{3}{c}{\textbf{GraphQuestions}}\\
    \cline{3-5}
    \cline{7-9}
     & & \textbf{Acc} & \textbf{Recall} & \textbf{EM}& & \textbf{Acc} & \textbf{Recall} & \textbf{EM}\\
    \midrule
    \multicolumn{9}{c}{\textit{ChatGPT as QA model}}\\
    \multirow{2}{*}{\textbf{None}} & No Knowledge & 28.91& 22.81& 20.14& &35.87 &25.76 &22.09 \\
    & Triple &57.76 & 49.67&44.73 & &55.03 &46.65 &41.63\\
    \cdashline{1-9}
    \multirow{4}{*}{\textbf{Llama-2}} & KG-to-Text &54.75 & 47.35&42.44 & &49.73 &40.00 &33.74 \\
    & Summary &58.14 &51.38 &46.38 & &\underline{52.94} &44.70 &38.41\\
     & CoTKR &\underline{58.64} &\underline{52.33} &\underline{47.88} & &51.36 &\underline{45.20} &\underline{39.96}\\
     & CoTKR+PA & \textbf{59.25} &\textbf{53.52} &\textbf{49.64} & &\textbf{56.78} &\textbf{47.99} &\textbf{42.46}\\
    \cdashline{1-9}
     \multirow{4}{*}{\textbf{Llama-3}} & KG-to-Text &55.76 &48.41 &43.90 & &52.40 &45.06 &39.83\\
     & Summary &57.55 &51.06 &46.80 & &\underline{54.95} &46.86 &40.75\\
     & CoTKR &\underline{58.33} &\underline{52.55} &\underline{48.65} & &53.19 &\underline{47.23} &\underline{43.17}\\
     & CoTKR+PA &\textbf{61.51} & \textbf{56.08}& \textbf{52.67} & &\textbf{56.37} &\textbf{49.31} & \textbf{45.26}\\
    \cdashline{1-9}
     \multirow{3}{*}{\textbf{ChatGPT}} & KG-to-Text &56.32 &49.05 &44.73 & &53.53 &45.59 &41.17\\
     & Summary &\underline{58.54} &\underline{51.81} &\underline{47.29} & &\textbf{55.62} &\textbf{48.93} &\textbf{44.97}\\
     & CoTKR &\textbf{59.87} &\textbf{53.19} &\textbf{49.02} & &\underline{54.28} &\underline{48.18} &\underline{44.68}\\
    \midrule
    \multicolumn{9}{c}{\textit{Mistral as QA model}}\\
     \multirow{2}{*}{\textbf{None}} & No Knowledge &29.44 &23.13 &20.30 & &38.20 &26.92 &22.13 \\
     & Triple &54.47 &47.78 &43.25 & & 51.32& 45.97&41.67\\
    \cdashline{1-9}
     \multirow{4}{*}{\textbf{Llama-2}} & KG-to-Text &49.49 &42.91 &38.41 & &44.59 & 37.98&32.82 \\
     & Summary &54.10 &47.79 &43.15 & &49.85 &42.33 &36.45\\
     & CoTKR &\underline{56.75} &\underline{51.10} & \underline{46.71} & &\underline{50.19} &\underline{43.73} &\underline{38.54}\\
     & CoTKR+PA &\textbf{58.15} & \textbf{52.98}&\textbf{49.13} & &\textbf{55.07} &\textbf{47.02} &\textbf{41.71}\\
    \cdashline{1-9}
     \multirow{4}{*}{\textbf{Llama-3}} & KG-to-Text &50.64 &44.32 &40.13 & &49.06 & 43.04&38.25\\
     & Summary &53.84 &47.71 &43.49 & &52.03 & 44.30&38.50\\
     & CoTKR &\underline{56.47} &\underline{51.33} &\underline{47.36} & & \underline{52.65}& \underline{46.48}&\underline{42.21}\\
     & CoTKR+PA &\textbf{59.31} &\textbf{54.13} &\textbf{50.24}& &\textbf{54.82} & \textbf{47.76}&\textbf{43.09}\\
    \cdashline{1-9}
     \multirow{3}{*}{\textbf{ChatGPT}} & KG-to-Text &51.04 &44.87 &40.97 & &49.14 &43.04 &38.83\\
     & Summary &\underline{54.44} & \underline{48.16} &\underline{43.97} & &\underline{52.28} &\underline{47.10} &\underline{43.30}\\
     & CoTKR &\textbf{57.28} &\textbf{51.14} &\textbf{47.09} & & \textbf{52.82}&\textbf{47.13} &\textbf{43.55}\\
    \bottomrule
\end{tabular}
}
\caption{The overall results of CoTKR and the baselines on GrailQA and GraphQuestions using 2-Hop as retrieval method. For each combination of a knowledge rewriter (KR) LLM and a QA model, the best and second-best results are highlighted in bold and underlined, respectively.}
\label{Table 1}
\end{table*}

To comprehensively evaluate various knowledge rewriting methods, we employ the widely used 2-Hop retrieval method. Table \ref{Table 1} presents the overall results. We observe that: \textbf{(1) Our method outperforms the baselines across most evaluation metrics and LLMs, confirming the effectiveness of our knowledge rewriting strategy.} This also demonstrates the broad practical applicability of \textsf{CoTKR}, effective for both open-source LLMs requiring fine-tuning and closed-source LLMs using ICL. Integrating question-related knowledge significantly improves QA performance compared with direct question answering, underscoring the efficacy of the RAG paradigm in KGQA. \textsf{KG-to-Text} exhibits the weakest performance, indicating that mere conversion of triples into text may result in loss of information inherent in the subgraph. \textsf{Summary} outperforms \textsf{KG-to-Text} but generally lags behind \textsf{CoTKR}/\textsf{CoTKR+PA}, suggesting that filtering out irrelevant knowledge is effective but not adequate. \textbf{(2) CoTKR+PA matches or even surpasses the performance of ChatGPT as the knowledge rewriter, proving the effectiveness of our training framework and the preference alignment.} \textsf{CoTKR+PA} outperforms \textsf{CoTKR}, indicating that preference alignment can bridge the preference gap between the knowledge rewriter and the QA model, thereby enhancing the quality of knowledge representation. \textbf{(3) A well-crafted knowledge representation is crucial for LLM used in KGQA.} Although \textsf{Triple} does not require an additional knowledge rewriting module, it provides a strong baseline and, in some cases, outperforms \textsf{KG-to-Text} and \textsf{Summary}. Conversely, \textsf{CoTKR+PA} consistently surpasses \textsf{Triple}. This indicates that \textsf{Triple} is simple yet effective and explains its widespread use in existing work. On the other hand, it demonstrates that a carefully designed knowledge representation can effectively enhance the performance of KGQA.

\subsection{Impact of Retrieval Methods}

\begin{figure}[h]
  \includegraphics[width=\columnwidth]{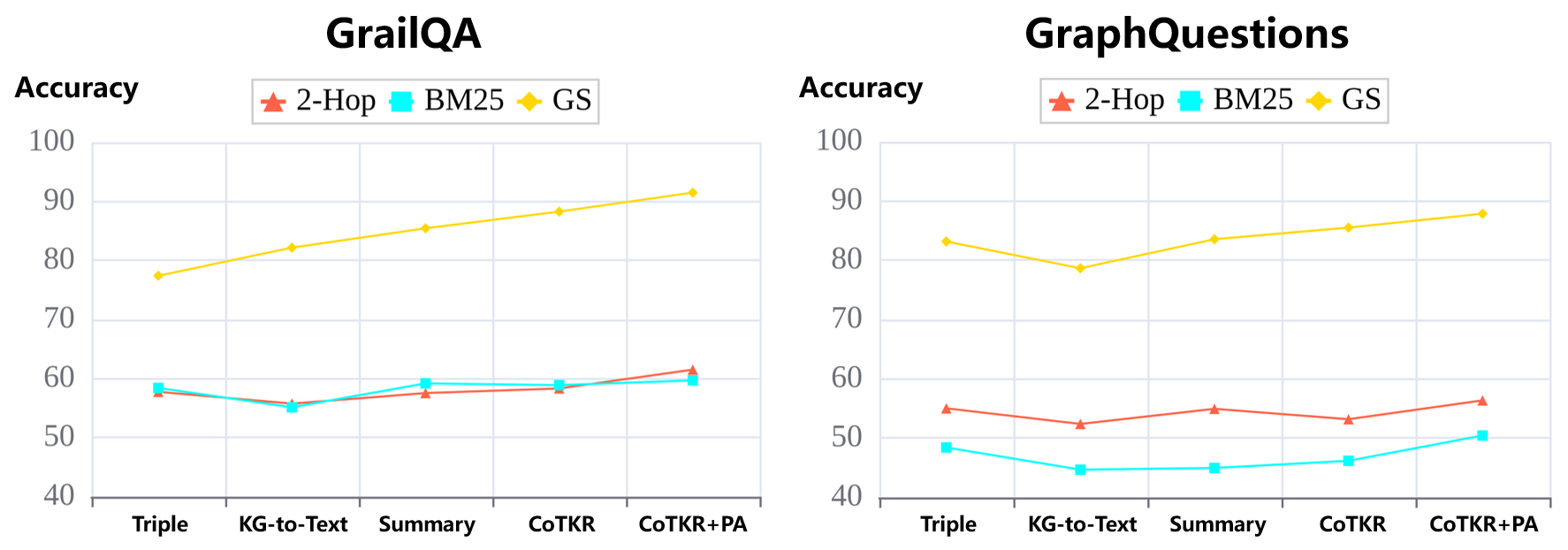}
  \caption{KGQA results using different knowledge rewriters and different retrieval methods.}
  \label{fig 4}
\end{figure}

To investigate the impact of retrieval methods, we select Llama-3 as the knowledge rewriter and ChatGPT as the QA model. According to the results shown in Figure \ref{fig 4}, we have the following observations: \textbf{(1) 2-Hop retrieval method may be insufficient for more challenging questions, but it is suitable for simpler ones.} Both BM25 and 2-Hop perform similarly on GrailQA, but 2-Hop shows a significant advantage over BM25 on GraphQuestions. This is likely because GrailQA is a more complex benchmark with a larger question-related subgraph, making 2-hop subgraphs often inadequate. Conversely, for GraphQuestions, a 2-hop subgraph usually provides precise context for most questions.  \textbf{(2) The design of a high-quality retriever remains an open problem.} GS significantly outperforms BM25 and 2-Hop, indicating that retrieval noise substantially affects KGQA performance. \textbf{(3) CoTKR consistently outperforms all baselines across various retrieval methods, demonstrating its robustness and practicality.}

\subsection{Comparison with Triple Method}
\begin{figure}[h]
  \includegraphics[width=\columnwidth]{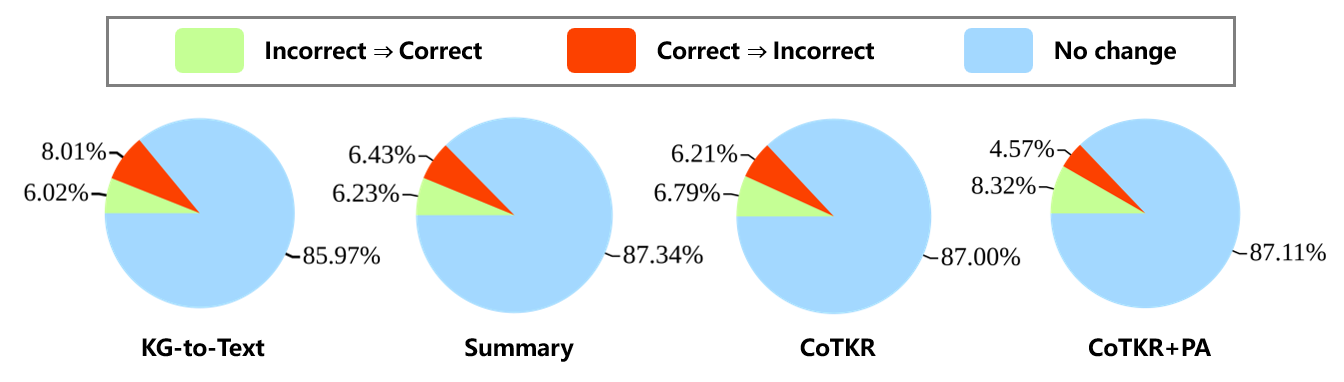}
  \caption{The comparative results on GrailQA. We use 2-Hop as the retrieval method.}
  \label{fig 5}
\end{figure}

Several studies \cite{DBLP:journals/corr/abs-2402-11541,baek-etal-2023-knowledge} suggest that LLMs can better comprehend triple-form text compared with natural language. However, our results show the contrary. Therefore, we delve deeper into this issue by comparing knowledge rewriting methods that use triple-form text as input (i.e., \textsf{KG-to-Text}, \textsf{Summary}, \textsf{CoTKR}, \textsf{CoTKR+PA}) with \textsf{Triple}. We use Accuracy as the criterion to evaluate the correctness of responses. For each method, we consider three scenarios: \textbf{(1) Incorrect$\rightarrow$Correct:} \textsf{Triple} provides a wrong answer, but the comparative method answers correctly. \textbf{(2) Correct->Incorrect:} \textsf{Triple} answers correctly, but the comparative method answers incorrectly. \textbf{(3) No change:} both \textsf{Triple} and the comparative method answer correctly or incorrectly. We adopt Llama-3 as the knowledge rewriter and ChatGPT as the QA model, with 2-Hop as the retrieval method. Then we calculate the proportions of three distinct cases within GrailQA. From the results shown in Figure \ref{fig 5}, we draw the following conclusions: \textbf{(1) KG-to-Text and Summary have a predominantly negative impact, partially validating the conclusions of prior studies.} \textsf{Triple} provides a strong baseline, and the adoption of \textsf{KG-to-Text} and \textsf{Summary} leads to more incorrect answers. This indicates that LLMs can understand triple-form text effectively, and using simple knowledge rewriting methods leads to loss of information. \textbf{(2) Well-designed knowledge representations substantially benefit the question-answering model.} The knowledge representations rewritten by \textsf{CoTKR/CoTKR+PA} generally enhance the QA model's performance. This reflects that the suboptimal knowledge representations in previous work are key contributors to performance degradation. Our method generates comprehensive and semantically coherent knowledge representations, thereby improving the efficacy of KGQA.

\subsection{Effectiveness of Data Augmentation}

\begin{table}[h]
\centering
\small
\resizebox{\columnwidth}{!}{
\begin{tabular}{llccccccc}
    \toprule
     \multirow{2}{*}{\textbf{KR LLMs}} & \multirow{2}{*}{\textbf{Methods}} & \multicolumn{3}{c}{\textbf{ChatGPT}} & &\multicolumn{3}{c}{\textbf{Mistral}}\\
    \cline{3-5}
    \cline{7-9}
     & & \textbf{Acc} & \textbf{Recall} & \textbf{EM}& & \textbf{Acc} & \textbf{Recall} & \textbf{EM}\\
    \midrule
    \multirow{3}{*}{\textbf{Llama-2}} & CoTKR & 51.36 &45.20 &39.96 & & 50.19&43.73 &38.54 \\
    & CoTKR+PA\textsuperscript{*} & 54.57 &47.40 &\textbf{42.88} & & 52.99&45.69 &41.09 \\
     & CoTKR+PA & \textbf{56.78} &\textbf{47.99} &42.46 & &\textbf{55.07} &\textbf{47.02} & \textbf{41.71}\\
    \cdashline{1-9}
     \multirow{3}{*}{\textbf{Llama-3}} & CoTKR & 53.19 &47.23 &43.17 & &52.65 &46.48 &42.21 \\
     & CoTKR+PA\textsuperscript{*} & 54.70 & 48.79&44.72 & &53.24 & 47.64&\textbf{43.63} \\
     & CoTKR+PA & \textbf{56.37}&\textbf{49.31} &\textbf{45.26} & &\textbf{54.82} &\textbf{47.76} &43.09 \\
    \bottomrule
\end{tabular}
}
\caption{Ablation study results for data augmentation on GraphQuestions, employing the 2-Hop retrieval method.}
\label{Table 2}
\end{table}

\begin{table*}[!h]
\centering
\small
\resizebox{\textwidth}{!}{
\begin{tabular}{lp{15cm}}
    \toprule
    \multicolumn{2}{l}{\textbf{Question}: what is the unit of area that the measurement system that have an electric field strength units of \textcolor{darkgreen}{volt per metre} have?}\\
    \multicolumn{2}{l}{\textbf{Answer}: \textcolor{orange}{Square kilometer}, \textcolor{orange}{Square meter}}\\
    \midrule
    \textbf{KR Methods}& \multicolumn{1}{c}{\textbf{KR Results}}\\
    \midrule
    \multirow{2}{*}{\textbf{Summary}} & The International System of Units includes various measurement systems for different physical quantities such as length, area, volume, conductance, resistance, current, force, and pressure. The measurement system with an electric field strength unit of \textcolor{darkgreen}{Volt per meter} does not directly relate to a specific area unit within the International System of Units.\\
    \cdashline{1-2}
    \multirow{6}{*}{\textbf{CoTKR}} & Reason: I need to know the unit of area associated with the measurement system that has an electric field strength unit of \textcolor{darkgreen}{Volt per meter}. \newline Summarize: The International System of Units includes various units for different physical quantities such as length, area, volume, etc.\newline Reason: I need to find the specific unit of area in the International System of Units that corresponds to the electric field strength unit of \textcolor{darkgreen}{Volt per meter}.\newline Summarize: The International System of Units includes units like \textcolor{orange}{Square kilometer}, \textcolor{orange}{Square meter}, etc. for measuring area.\\
    \cdashline{1-2}
    \multirow{4}{*}{\textbf{CoTKR+PA}}& Reason: I need to know which measurement system includes the unit for electric field strength in \textcolor{darkgreen}{Volt per meter}.\newline Knowledge: The International System of Units contains the unit for electric field strength in \textcolor{darkgreen}{Volt per meter}.\newline Reason: I need to identify the unit of area within the International System of Units.\newline Knowledge: The International System of Units includes the units \textcolor{orange}{Square kilometer} and \textcolor{orange}{Square meter} as area units.\\
    \bottomrule
\end{tabular}
}
\caption{An example of knowledge rewriting results for different methods.}
\label{Table 3}
\end{table*}

To evaluate the effectiveness of data augmentation, we compare the performance of three variants: \textsf{CoTKR}, \textsf{CoTKR+PA}, and \textsf{CoTKR+PA}\textsuperscript{*} (using supervised fine-tuning and preference alignment without data augmentation). The experimental results are shown in Table \ref{Table 2}, from which we can draw two conclusions: \textbf{(1) CoTKR+PA\textsuperscript{*} generally outperforms CoTKR+PA, indicating that PAQAF does not solely rely on data augmentation based on ChatGPT.} \textbf{(2) CoTKR+PA performs best in most scenarios, proving that data augmentation enhances the preference alignment.}

\subsection{Case Study}

In this section, we compare \textsf{Summary} with \textsf{CoTKR} through an example. (Please refer to Appendix \ref{case} for the full comparison result.) As illustrated in Table \ref{Table 3}, \textsf{Summary} struggles to extract useful information when faced with an abundance of triples. In contrast, \textsf{CoTKR}, leveraging CoT reasoning, effectively emphasizes the key evidence (i.e., Square meter) in the second rewriting step. Furthermore, after preference alignment, \textsf{CoTKR+PA} is capable of generating more natural reasoning steps, significantly enhancing its applicability to KGQA.

\section{Conclusion}
In this paper, we propose \textbf{C}hain-\textbf{o}f-\textbf{T}hought Enhanced \textbf{K}nowledge \textbf{R}ewriting, CoTKR, for higher quality knowledge representation of triples in KG augmented QA. To bridge the preference gap between the knowledge rewriter and the QA model, we propose \textbf{P}reference \textbf{A}lignment from \textbf{Q}uestion \textbf{A}nswering \textbf{F}eedback, PAQAF. Experimental results demonstrate that, compared with existing knowledge rewriting methods, CoTKR can generate the most beneficial knowledge representation for QA models. In future work, we will go beyond KGQA to explore knowledge representations for other kinds of structured data for RAG.


\section{Limitations}
We acknowledge the limitations of this work. (1) This study is limited to KGQA and does not explore broader application scenarios. Therefore, we did not design experiments to explore whether CoTKR is effective for all or most RAG scenarios. In future work, we aim to expand the range of data sources. We intend to design a knowledge rewriting method that can be applied to not only KGs but also tables, textual data, and other formats. This enhancement will allow the QA framework to access knowledge from a wider range of sources, thus improving its practicality. Furthermore, we plan to investigate a knowledge representation beneficial for various downstream tasks, such as fact verification and dialogue generation. (2) The training framework for CoTKR depends on the powerful capabilities of closed-source LLMs. However, these models have inherent limitations, and the training data they generate contains noise, which constrains the performance ceiling of CoTKR.

\section{Ethical Considerations}
We explore optimizing knowledge representations for KGQA on public benchmarks, avoiding any potential harm to any individuals or groups. To promote transparency and facilitate replication of our research, we provide the technical details necessary for reproducing our results and release both the source code and the collected data. Our code and data are available for academic research, commercial use, and other applications.

It is important to acknowledge the potential risks and ethical considerations associated with LLMs. In this study, we construct the training data using ChatGPT and implement our knowledge rewriters based on LLMs. Due to the inherent limitations of LLMs, including factual inaccuracies, racial discrimination, and gender bias, our knowledge rewriters might generate incorrect content or inadvertently reflect prevalent societal biases.

\section*{Acknowledgments}
This work is partially supported by National Nature Science Foundation of China under No. U21A20488, by the project "Key Laboratory of rich-media Digital Publishing Content Organization and Knowledge Service Open Fund-Research on Knowledge-enhanced Training Techinques of Large Language Model" No. ZD2024-04/01, by the EPSRC project OntoEm (EP/Y017706/1), and by Southeast University-China Mobile Research Institute Joint Innovation Center. It is funded by Southeast University-China Mobile Research Institute Joint Innovation Center. We thank the Big Data Computing Center of Southeast University for providing the facility support on the numerical calculations in this paper. ChatGPT was used to enhance the readability of some of the text and improve the language of this paper, after the content was first added manually. All material was then checked manually before submission.


\bibliography{custom}

@article{DBLP:journals/corr/abs-2205-01068,
  author       = {Susan Zhang and
                  Stephen Roller and
                  Naman Goyal and
                  Mikel Artetxe and
                  Moya Chen and
                  Shuohui Chen and
                  Christopher Dewan and
                  Mona T. Diab and
                  Xian Li and
                  Xi Victoria Lin and
                  Todor Mihaylov and
                  Myle Ott and
                  Sam Shleifer and
                  Kurt Shuster and
                  Daniel Simig and
                  Punit Singh Koura and
                  Anjali Sridhar and
                  Tianlu Wang and
                  Luke Zettlemoyer},
  title        = {{OPT:} Open Pre-trained Transformer Language Models},
  journal      = {CoRR},
  volume       = {abs/2205.01068},
  year         = {2022}
}

@inproceedings{DBLP:conf/iclr/SanhWRBSACSRDBX22,
  author       = {Victor Sanh and
                  Albert Webson and
                  Colin Raffel and
                  Stephen H. Bach and
                  Lintang Sutawika and
                  Zaid Alyafeai and
                  Antoine Chaffin and
                  Arnaud Stiegler and
                  Arun Raja and
                  Manan Dey and
                  M Saiful Bari and
                  Canwen Xu and
                  Urmish Thakker and
                  Shanya Sharma Sharma and
                  Eliza Szczechla and
                  Taewoon Kim and
                  Gunjan Chhablani and
                  Nihal V. Nayak and
                  Debajyoti Datta and
                  Jonathan Chang and
                  Mike Tian{-}Jian Jiang and
                  Han Wang and
                  Matteo Manica and
                  Sheng Shen and
                  Zheng Xin Yong and
                  Harshit Pandey and
                  Rachel Bawden and
                  Thomas Wang and
                  Trishala Neeraj and
                  Jos Rozen and
                  Abheesht Sharma and
                  Andrea Santilli and
                  Thibault F{\'{e}}vry and
                  Jason Alan Fries and
                  Ryan Teehan and
                  Teven Le Scao and
                  Stella Biderman and
                  Leo Gao and
                  Thomas Wolf and
                  Alexander M. Rush},
  title        = {Multitask Prompted Training Enables Zero-Shot Task Generalization},
  booktitle    = {{ICLR}},
  publisher    = {OpenReview.net},
  year         = {2022}
}

@inproceedings{DBLP:conf/nips/BrownMRSKDNSSAA20,
  author       = {Tom B. Brown and
                  Benjamin Mann and
                  Nick Ryder and
                  Melanie Subbiah and
                  Jared Kaplan and
                  Prafulla Dhariwal and
                  Arvind Neelakantan and
                  Pranav Shyam and
                  Girish Sastry and
                  Amanda Askell and
                  Sandhini Agarwal and
                  Ariel Herbert{-}Voss and
                  Gretchen Krueger and
                  Tom Henighan and
                  Rewon Child and
                  Aditya Ramesh and
                  Daniel M. Ziegler and
                  Jeffrey Wu and
                  Clemens Winter and
                  Christopher Hesse and
                  Mark Chen and
                  Eric Sigler and
                  Mateusz Litwin and
                  Scott Gray and
                  Benjamin Chess and
                  Jack Clark and
                  Christopher Berner and
                  Sam McCandlish and
                  Alec Radford and
                  Ilya Sutskever and
                  Dario Amodei},
  title        = {Language Models are Few-Shot Learners},
  booktitle    = {NeurIPS},
  year         = {2020}
}

@article{PRKSC2023,
    author = {Jeff Z. Pan and Simon Razniewski and Jan-Christoph Kalo and Sneha Singhania and Jiaoyan Chen and Stefan Dietze and Hajira Jabeen and Janna Omeliyanenko and Wen Zhang and Matteo Lissandrini and ussa Biswas and Gerard de Melo and Angela Bonifati and Edlira Vakaj and Mauro Dragoni and amien Graux},
    title = {Large Language Models and Knowledge Graphs: Opportunities and Challenges},
    journal = {Transactions on Graph Data and Knowledge},
    year = {2023}
}

@book{PVGW2017,
 editor    = {Pan, J.Z. and Vetere, G. and Gomez-Perez, J.M. and Wu, H.},
 title     = {{Exploiting Linked Data and Knowledge Graphs for Large Organisations}},
 publisher   ={Springer},
 isbn      = {978-3-319-45652-2}, 
 year       ={2017}
}

@book{PCEH+2017,
 editor    = {Pan, J.Z. and Calvanese, D. and Eiter, T. and Horrocks, I. and Kifer, M. and Lin, F. and Zhao, Y.},
 title     = {{Reasoning Web: Logical Foundation of Knowledge Graph Construction and Querying Answering}},
 publisher   ={Springer},
 isbn      = {978-3-319-49493-7}, 
 year       ={2017}
}

@inproceedings{HLVPP2023, 
  author = {Huang, Wenyu and Lapata, Mirella and Vougiouklis, Pavlos and Papasarantopoulos, Nikos and Pan, Jeff Z.}, 
  booktitle = {Proceedings of the 13th International Joint Conference on Natural Language Processing and the 3rd Conference of
the Asia-Pacific Chapter of the Association for Computational Linguistics}, 
  pages = {1012-1025}, title = {Retrieval Augmented Generation with Rich Answer Encoding.}, 
  year = 2023
}

@InProceedings{HUGP2023,
 author    = {Jie He and Simon Chi Lok U and Víctor Gutiérrez-Basulto and Jeff Z. Pan},
 title     = {{BUCA: A Binary Classification Approach to Unsupervised Commonsense
Question Answering}},
 booktitle = {Proceedings of the 61st Annual Meeting of the Association for Computational Linguistics (ACL 2023)},
 year      = {2023}
 }

@inproceedings{DBLP:conf/iclr/WeiBZGYLDDL22,
  author       = {Jason Wei and
                  Maarten Bosma and
                  Vincent Y. Zhao and
                  Kelvin Guu and
                  Adams Wei Yu and
                  Brian Lester and
                  Nan Du and
                  Andrew M. Dai and
                  Quoc V. Le},
  title        = {Finetuned Language Models are Zero-Shot Learners},
  booktitle    = {{ICLR}},
  publisher    = {OpenReview.net},
  year         = {2022}
}

@inproceedings{DBLP:conf/nips/KojimaGRMI22,
  author       = {Takeshi Kojima and
                  Shixiang Shane Gu and
                  Machel Reid and
                  Yutaka Matsuo and
                  Yusuke Iwasawa},
  title        = {Large Language Models are Zero-Shot Reasoners},
  booktitle    = {NeurIPS},
  year         = {2022}
}

@article{DBLP:journals/www/HuWQMCPA23,
  author       = {Nan Hu and
                  Yike Wu and
                  Guilin Qi and
                  Dehai Min and
                  Jiaoyan Chen and
                  Jeff Z. Pan and
                  Zafar Ali},
  title        = {An empirical study of pre-trained language models in simple knowledge
                  graph question answering},
  journal      = {World Wide Web {(WWW)}},
  volume       = {26},
  number       = {5},
  pages        = {2855--2886},
  year         = {2023}
}

@article{DBLP:journals/corr/abs-2303-07992,
  author       = {Yiming Tan and
                  Dehai Min and
                  Yu Li and
                  Wenbo Li and
                  Nan Hu and
                  Yongrui Chen and
                  Guilin Qi},
  title        = {Evaluation of ChatGPT as a Question Answering System for Answering
                  Complex Questions},
  journal      = {CoRR},
  volume       = {abs/2303.07992},
  year         = {2023}
}

@inproceedings{DBLP:conf/emnlp/MaGHZD23,
  author       = {Xinbei Ma and
                  Yeyun Gong and
                  Pengcheng He and
                  Hai Zhao and
                  Nan Duan},
  title        = {Query Rewriting in Retrieval-Augmented Large Language Models},
  booktitle    = {{EMNLP}},
  pages        = {5303--5315},
  publisher    = {Association for Computational Linguistics},
  year         = {2023}
}

@inproceedings{DBLP:conf/acl/TrivediBKS23,
  author       = {Harsh Trivedi and
                  Niranjan Balasubramanian and
                  Tushar Khot and
                  Ashish Sabharwal},
  title        = {Interleaving Retrieval with Chain-of-Thought Reasoning for Knowledge-Intensive
                  Multi-Step Questions},
  booktitle    = {{ACL} {(1)}},
  pages        = {10014--10037},
  publisher    = {Association for Computational Linguistics},
  year         = {2023}
}

@article{DBLP:journals/corr/abs-2309-11206,
  author       = {Yike Wu and
                  Nan Hu and
                  Sheng Bi and
                  Guilin Qi and
                  Jie Ren and
                  Anhuan Xie and
                  Wei Song},
  title        = {Retrieve-Rewrite-Answer: {A} KG-to-Text Enhanced LLMs Framework for
                  Knowledge Graph Question Answering},
  journal      = {CoRR},
  volume       = {abs/2309.11206},
  year         = {2023}
}

@inproceedings{DBLP:conf/nips/LewisPPPKGKLYR020,
  author       = {Patrick S. H. Lewis and
                  Ethan Perez and
                  Aleksandra Piktus and
                  Fabio Petroni and
                  Vladimir Karpukhin and
                  Naman Goyal and
                  Heinrich K{\"{u}}ttler and
                  Mike Lewis and
                  Wen{-}tau Yih and
                  Tim Rockt{\"{a}}schel and
                  Sebastian Riedel and
                  Douwe Kiela},
  title        = {Retrieval-Augmented Generation for Knowledge-Intensive {NLP} Tasks},
  booktitle    = {NeurIPS},
  year         = {2020}
}

@article{DBLP:journals/corr/abs-2312-10997,
  author       = {Yunfan Gao and
                  Yun Xiong and
                  Xinyu Gao and
                  Kangxiang Jia and
                  Jinliu Pan and
                  Yuxi Bi and
                  Yi Dai and
                  Jiawei Sun and
                  Qianyu Guo and
                  Meng Wang and
                  Haofen Wang},
  title        = {Retrieval-Augmented Generation for Large Language Models: {A} Survey},
  journal      = {CoRR},
  volume       = {abs/2312.10997},
  year         = {2023}
}

@inproceedings{baek-etal-2023-knowledge,
    title = "Knowledge-Augmented Language Model Prompting for Zero-Shot Knowledge Graph Question Answering",
    author = "Baek, Jinheon  and
      Aji, Alham Fikri  and
      Saffari, Amir",
    editor = "Dalvi Mishra, Bhavana  and
      Durrett, Greg  and
      Jansen, Peter  and
      Neves Ribeiro, Danilo  and
      Wei, Jason",
    booktitle = "Proceedings of the 1st Workshop on Natural Language Reasoning and Structured Explanations (NLRSE)",
    month = jun,
    year = "2023",
    address = "Toronto, Canada",
    publisher = "Association for Computational Linguistics",
    url = "https://aclanthology.org/2023.nlrse-1.7",
    doi = "10.18653/v1/2023.nlrse-1.7",
    pages = "78--106",
}

@inproceedings{sen-etal-2023-knowledge,
    title = "Knowledge Graph-augmented Language Models for Complex Question Answering",
    author = "Sen, Priyanka  and
      Mavadia, Sandeep  and
      Saffari, Amir",
    editor = "Dalvi Mishra, Bhavana  and
      Durrett, Greg  and
      Jansen, Peter  and
      Neves Ribeiro, Danilo  and
      Wei, Jason",
    booktitle = "Proceedings of the 1st Workshop on Natural Language Reasoning and Structured Explanations (NLRSE)",
    month = jun,
    year = "2023",
    address = "Toronto, Canada",
    publisher = "Association for Computational Linguistics",
    url = "https://aclanthology.org/2023.nlrse-1.1",
    doi = "10.18653/v1/2023.nlrse-1.1",
    pages = "1--8",
}

@article{DBLP:journals/corr/abs-2401-00426,
  author       = {Chaojie Wang and
                  Yishi Xu and
                  Zhong Peng and
                  Chenxi Zhang and
                  Bo Chen and
                  Xinrun Wang and
                  Lei Feng and
                  Bo An},
  title        = {keqing: knowledge-based question answering is a nature chain-of-thought
                  mentor of {LLM}},
  journal      = {CoRR},
  volume       = {abs/2401.00426},
  year         = {2024}
}

@article{DBLP:journals/corr/abs-2403-02966,
  author       = {SungHo Ko and
                  Hyunjin Cho and
                  Hyungjoo Chae and
                  Jinyoung Yeo and
                  Dongha Lee},
  title        = {Evidence-Focused Fact Summarization for Knowledge-Augmented Zero-Shot
                  Question Answering},
  journal      = {CoRR},
  volume       = {abs/2403.02966},
  year         = {2024}
}

@inproceedings{DBLP:conf/nips/Wei0SBIXCLZ22,
  author       = {Jason Wei and
                  Xuezhi Wang and
                  Dale Schuurmans and
                  Maarten Bosma and
                  Brian Ichter and
                  Fei Xia and
                  Ed H. Chi and
                  Quoc V. Le and
                  Denny Zhou},
  title        = {Chain-of-Thought Prompting Elicits Reasoning in Large Language Models},
  booktitle    = {NeurIPS},
  year         = {2022}
}

@inproceedings{DBLP:conf/iclr/YaoZYDSN023,
  author       = {Shunyu Yao and
                  Jeffrey Zhao and
                  Dian Yu and
                  Nan Du and
                  Izhak Shafran and
                  Karthik R. Narasimhan and
                  Yuan Cao},
  title        = {ReAct: Synergizing Reasoning and Acting in Language Models},
  booktitle    = {{ICLR}},
  publisher    = {OpenReview.net},
  year         = {2023}
}

@inproceedings{DBLP:conf/nips/RafailovSMMEF23,
  author       = {Rafael Rafailov and
                  Archit Sharma and
                  Eric Mitchell and
                  Christopher D. Manning and
                  Stefano Ermon and
                  Chelsea Finn},
  title        = {Direct Preference Optimization: Your Language Model is Secretly a
                  Reward Model},
  booktitle    = {NeurIPS},
  year         = {2023}
}

@article{DBLP:journals/corr/abs-2307-14192,
  author       = {Xiaodong Wu and
                  Ran Duan and
                  Jianbing Ni},
  title        = {Unveiling Security, Privacy, and Ethical Concerns of ChatGPT},
  journal      = {CoRR},
  volume       = {abs/2307.14192},
  year         = {2023}
}

@article{ray2023chatgpt,
  title={ChatGPT: A comprehensive review on background, applications, key challenges, bias, ethics, limitations and future scope. Internet of Things and Cyber-Physical Systems, 3, 121--154},
  author={Ray, Partha Pratim},
  journal={URL https://doi. org/10.1016/j. iotcps},
  volume={3},
  year={2023}
}

@article{DBLP:journals/corr/abs-2310-19852,
  author       = {Jiaming Ji and
                  Tianyi Qiu and
                  Boyuan Chen and
                  Borong Zhang and
                  Hantao Lou and
                  Kaile Wang and
                  Yawen Duan and
                  Zhonghao He and
                  Jiayi Zhou and
                  Zhaowei Zhang and
                  Fanzhi Zeng and
                  Kwan Yee Ng and
                  Juntao Dai and
                  Xuehai Pan and
                  Aidan O'Gara and
                  Yingshan Lei and
                  Hua Xu and
                  Brian Tse and
                  Jie Fu and
                  Stephen McAleer and
                  Yaodong Yang and
                  Yizhou Wang and
                  Song{-}Chun Zhu and
                  Yike Guo and
                  Wen Gao},
  title        = {{AI} Alignment: {A} Comprehensive Survey},
  journal      = {CoRR},
  volume       = {abs/2310.19852},
  year         = {2023}
}

@article{DBLP:journals/corr/abs-2307-12966,
  author       = {Yufei Wang and
                  Wanjun Zhong and
                  Liangyou Li and
                  Fei Mi and
                  Xingshan Zeng and
                  Wenyong Huang and
                  Lifeng Shang and
                  Xin Jiang and
                  Qun Liu},
  title        = {Aligning Large Language Models with Human: {A} Survey},
  journal      = {CoRR},
  volume       = {abs/2307.12966},
  year         = {2023}
}

@article{DBLP:journals/corr/abs-2311-06503,
  author       = {Yichi Zhang and
                  Zhuo Chen and
                  Yin Fang and
                  Lei Cheng and
                  Yanxi Lu and
                  Fangming Li and
                  Wen Zhang and
                  Huajun Chen},
  title        = {Knowledgeable Preference Alignment for LLMs in Domain-specific Question
                  Answering},
  journal      = {CoRR},
  volume       = {abs/2311.06503},
  year         = {2023}
}

@article{DBLP:journals/corr/abs-2401-06954,
  author       = {Zixuan Ke and
                  Weize Kong and
                  Cheng Li and
                  Mingyang Zhang and
                  Qiaozhu Mei and
                  Michael Bendersky},
  title        = {Bridging the Preference Gap between Retrievers and LLMs},
  journal      = {CoRR},
  volume       = {abs/2401.06954},
  year         = {2024}
}

@inproceedings{ma-etal-2023-query,
    title = "Query Rewriting in Retrieval-Augmented Large Language Models",
    author = "Ma, Xinbei  and
      Gong, Yeyun  and
      He, Pengcheng  and
      Zhao, Hai  and
      Duan, Nan",
    editor = "Bouamor, Houda  and
      Pino, Juan  and
      Bali, Kalika",
    booktitle = "Proceedings of the 2023 Conference on Empirical Methods in Natural Language Processing",
    month = dec,
    year = "2023",
    address = "Singapore",
    publisher = "Association for Computational Linguistics",
    url = "https://aclanthology.org/2023.emnlp-main.322",
    doi = "10.18653/v1/2023.emnlp-main.322",
    pages = "5303--5315",
}

@inproceedings{DBLP:conf/www/GuKVSLY021,
  author       = {Yu Gu and
                  Sue Kase and
                  Michelle Vanni and
                  Brian M. Sadler and
                  Percy Liang and
                  Xifeng Yan and
                  Yu Su},
  title        = {Beyond {I.I.D.:} Three Levels of Generalization for Question Answering
                  on Knowledge Bases},
  booktitle    = {{WWW}},
  pages        = {3477--3488},
  publisher    = {{ACM} / {IW3C2}},
  year         = {2021}
}

@inproceedings{DBLP:conf/emnlp/SuSSSGYY16,
  author       = {Yu Su and
                  Huan Sun and
                  Brian M. Sadler and
                  Mudhakar Srivatsa and
                  Izzeddin Gur and
                  Zenghui Yan and
                  Xifeng Yan},
  title        = {On Generating Characteristic-rich Question Sets for {QA} Evaluation},
  booktitle    = {{EMNLP}},
  pages        = {562--572},
  publisher    = {The Association for Computational Linguistics},
  year         = {2016}
}

@article{DBLP:journals/corr/abs-2302-13971,
  author       = {Hugo Touvron and
                  Thibaut Lavril and
                  Gautier Izacard and
                  Xavier Martinet and
                  Marie{-}Anne Lachaux and
                  Timoth{\'{e}}e Lacroix and
                  Baptiste Rozi{\`{e}}re and
                  Naman Goyal and
                  Eric Hambro and
                  Faisal Azhar and
                  Aur{\'{e}}lien Rodriguez and
                  Armand Joulin and
                  Edouard Grave and
                  Guillaume Lample},
  title        = {LLaMA: Open and Efficient Foundation Language Models},
  journal      = {CoRR},
  volume       = {abs/2302.13971},
  year         = {2023}
}

@article{DBLP:journals/ftir/RobertsonZ09,
  author       = {Stephen E. Robertson and
                  Hugo Zaragoza},
  title        = {The Probabilistic Relevance Framework: {BM25} and Beyond},
  journal      = {Found. Trends Inf. Retr.},
  volume       = {3},
  number       = {4},
  pages        = {333--389},
  year         = {2009}
}

@article{DBLP:journals/corr/abs-2308-13259,
  author       = {Keheng Wang and
                  Feiyu Duan and
                  Sirui Wang and
                  Peiguang Li and
                  Yunsen Xian and
                  Chuantao Yin and
                  Wenge Rong and
                  Zhang Xiong},
  title        = {Knowledge-Driven CoT: Exploring Faithful Reasoning in LLMs for Knowledge-intensive
                  Question Answering},
  journal      = {CoRR},
  volume       = {abs/2308.13259},
  year         = {2023}
}

@inproceedings{DBLP:conf/iclr/YuZNZL0HWWX23,
  author       = {Donghan Yu and
                  Sheng Zhang and
                  Patrick Ng and
                  Henghui Zhu and
                  Alexander Hanbo Li and
                  Jun Wang and
                  Yiqun Hu and
                  William Yang Wang and
                  Zhiguo Wang and
                  Bing Xiang},
  title        = {DecAF: Joint Decoding of Answers and Logical Forms for Question Answering
                  over Knowledge Bases},
  booktitle    = {{ICLR}},
  publisher    = {OpenReview.net},
  year         = {2023}
}

@inproceedings{DBLP:conf/sigir/LinMLYPN21,
  author       = {Jimmy Lin and
                  Xueguang Ma and
                  Sheng{-}Chieh Lin and
                  Jheng{-}Hong Yang and
                  Ronak Pradeep and
                  Rodrigo Frassetto Nogueira},
  title        = {Pyserini: {A} Python Toolkit for Reproducible Information Retrieval
                  Research with Sparse and Dense Representations},
  booktitle    = {{SIGIR}},
  pages        = {2356--2362},
  publisher    = {{ACM}},
  year         = {2021}
}

@inproceedings{DBLP:conf/aaai/BollackerCT07,
  author       = {Kurt D. Bollacker and
                  Robert P. Cook and
                  Patrick Tufts},
  title        = {Freebase: {A} Shared Database of Structured General Human Knowledge},
  booktitle    = {{AAAI}},
  pages        = {1962--1963},
  publisher    = {{AAAI} Press},
  year         = {2007}
}

@inproceedings{DBLP:conf/sigmod/BollackerEPST08,
  author       = {Kurt D. Bollacker and
                  Colin Evans and
                  Praveen K. Paritosh and
                  Tim Sturge and
                  Jamie Taylor},
  title        = {Freebase: a collaboratively created graph database for structuring
                  human knowledge},
  booktitle    = {{SIGMOD} Conference},
  pages        = {1247--1250},
  publisher    = {{ACM}},
  year         = {2008}
}

@inproceedings{DBLP:conf/iclr/HuSWALWWC22,
  author       = {Edward J. Hu and
                  Yelong Shen and
                  Phillip Wallis and
                  Zeyuan Allen{-}Zhu and
                  Yuanzhi Li and
                  Shean Wang and
                  Lu Wang and
                  Weizhu Chen},
  title        = {LoRA: Low-Rank Adaptation of Large Language Models},
  booktitle    = {{ICLR}},
  publisher    = {OpenReview.net},
  year         = {2022}
}

@inproceedings{DBLP:conf/www/DingLLQ24,
  author       = {Wentao Ding and
                  Jinmao Li and
                  Liangchuan Luo and
                  Yuzhong Qu},
  title        = {Enhancing Complex Question Answering over Knowledge Graphs through
                  Evidence Pattern Retrieval},
  booktitle    = {{WWW}},
  pages        = {2106--2115},
  publisher    = {{ACM}},
  year         = {2024}
}

@article{DBLP:journals/corr/abs-2402-11541,
  author       = {Xinbang Dai and
                  Yuncheng Hua and
                  Tongtong Wu and
                  Yang Sheng and
                  Guilin Qi},
  title        = {Counter-intuitive: Large Language Models Can Better Understand Knowledge
                  Graphs Than We Thought},
  journal      = {CoRR},
  volume       = {abs/2402.11541},
  year         = {2024}
}

@article{DBLP:journals/csur/JiLFYSXIBMF23,
  author       = {Ziwei Ji and
                  Nayeon Lee and
                  Rita Frieske and
                  Tiezheng Yu and
                  Dan Su and
                  Yan Xu and
                  Etsuko Ishii and
                  Yejin Bang and
                  Andrea Madotto and
                  Pascale Fung},
  title        = {Survey of Hallucination in Natural Language Generation},
  journal      = {{ACM} Comput. Surv.},
  volume       = {55},
  number       = {12},
  pages        = {248:1--248:38},
  year         = {2023}
}

@inproceedings{DBLP:conf/aaai/BianH0021,
  author       = {Ning Bian and
                  Xianpei Han and
                  Bo Chen and
                  Le Sun},
  title        = {Benchmarking Knowledge-Enhanced Commonsense Question Answering via
                  Knowledge-to-Text Transformation},
  booktitle    = {{AAAI}},
  pages        = {12574--12582},
  publisher    = {{AAAI} Press},
  year         = {2021}
}

@inproceedings{DBLP:conf/jist/0007HCGFP0Z22,
  author       = {Zhuo Chen and
                  Yufeng Huang and
                  Jiaoyan Chen and
                  Yuxia Geng and
                  Yin Fang and
                  Jeff Z. Pan and
                  Ningyu Zhang and
                  Wen Zhang},
  title        = {LaKo: Knowledge-driven Visual Question Answering via Late Knowledge-to-Text
                  Injection},
  booktitle    = {{IJCKG}},
  pages        = {20--29},
  publisher    = {{ACM}},
  year         = {2022}
}

@inproceedings{DBLP:conf/aaaiss/DernbachAZHC24,
  author       = {Stefan Dernbach and
                  Khushbu Agarwal and
                  Alejandro Zuniga and
                  Michael Henry and
                  Sutanay Choudhury},
  title        = {GLaM: Fine-Tuning Large Language Models for Domain Knowledge Graph
                  Alignment via Neighborhood Partitioning and Generative Subgraph Encoding},
  booktitle    = {{AAAI} Spring Symposia},
  pages        = {82--89},
  publisher    = {{AAAI} Press},
  year         = {2024}
}

@inproceedings{DBLP:conf/acl/SaxenaTT20,
  author       = {Apoorv Saxena and
                  Aditay Tripathi and
                  Partha P. Talukdar},
  title        = {Improving Multi-hop Question Answering over Knowledge Graphs using
                  Knowledge Base Embeddings},
  booktitle    = {{ACL}},
  pages        = {4498--4507},
  publisher    = {Association for Computational Linguistics},
  year         = {2020}
}

@inproceedings{DBLP:conf/iclr/JiangZ0W23,
  author       = {Jinhao Jiang and
                  Kun Zhou and
                  Xin Zhao and
                  Ji{-}Rong Wen},
  title        = {UniKGQA: Unified Retrieval and Reasoning for Solving Multi-hop Question
                  Answering Over Knowledge Graph},
  booktitle    = {{ICLR}},
  publisher    = {OpenReview.net},
  year         = {2023}
}

@article{touvron2023llama2,
  title={Llama 2: Open foundation and fine-tuned chat models},
  author={Touvron, Hugo and Martin, Louis and Stone, Kevin and Albert, Peter and Almahairi, Amjad and Babaei, Yasmine and Bashlykov, Nikolay and Batra, Soumya and Bhargava, Prajjwal and Bhosale, Shruti and others},
  journal={arXiv preprint arXiv:2307.09288},
  year={2023}
}

@article{llama3modelcard,
title={Llama 3 Model Card},
author={AI@Meta},
year={2024},
url = {https://github.com/meta-llama/llama3/blob/main/MODEL_CARD.md}
}

@article{jiang2023mistral,
  title={Mistral 7B},
  author={Jiang, Albert Q and Sablayrolles, Alexandre and Mensch, Arthur and Bamford, Chris and Chaplot, Devendra Singh and Casas, Diego de las and Bressand, Florian and Lengyel, Gianna and Lample, Guillaume and Saulnier, Lucile and others},
  journal={arXiv preprint arXiv:2310.06825},
  year={2023}
}

@article{DBLP:journals/dint/AzariaAR24,
  author       = {Amos Azaria and
                  Rina Azoulay and
                  Shulamit Reches},
  title        = {ChatGPT is a Remarkable Tool - For Experts},
  journal      = {Data Intell.},
  volume       = {6},
  number       = {1},
  pages        = {240--296},
  year         = {2024}
}

@article{publisher/Beijing,
  author = "Huajun Chen",
  title = "Large Knowledge Model: Perspectives and Challenges",
  journal = "Data Intelligence",
  year = "2024",
  volume = "6",
  number = "3",
  pages = "587-620"
}

@article{DBLP:journals/dint/LiZLYC23,
  author       = {Linhan Li and
                  Huaping Zhang and
                  Chunjin Li and
                  Haowen You and
                  Wenyao Cui},
  title        = {Evaluation on ChatGPT for Chinese Language Understanding},
  journal      = {Data Intell.},
  volume       = {5},
  number       = {4},
  pages        = {885--903},
  year         = {2023}
}

@inproceedings{DBLP:conf/naacl/TalmorB18,
  author       = {Alon Talmor and
                  Jonathan Berant},
  title        = {The Web as a Knowledge-Base for Answering Complex Questions},
  booktitle    = {{NAACL-HLT}},
  pages        = {641--651},
  publisher    = {Association for Computational Linguistics},
  year         = {2018}
}

@inproceedings{DBLP:conf/emnlp/SottanaLZY23,
  author       = {Andrea Sottana and
                  Bin Liang and
                  Kai Zou and
                  Zheng Yuan},
  title        = {Evaluation Metrics in the Era of {GPT-4:} Reliably Evaluating Large
                  Language Models on Sequence to Sequence Tasks},
  booktitle    = {{EMNLP}},
  pages        = {8776--8788},
  publisher    = {Association for Computational Linguistics},
  year         = {2023}
}

@inproceedings{DBLP:conf/emnlp/LiuIXWXZ23,
  author       = {Yang Liu and
                  Dan Iter and
                  Yichong Xu and
                  Shuohang Wang and
                  Ruochen Xu and
                  Chenguang Zhu},
  title        = {G-Eval: {NLG} Evaluation using Gpt-4 with Better Human Alignment},
  booktitle    = {{EMNLP}},
  pages        = {2511--2522},
  publisher    = {Association for Computational Linguistics},
  year         = {2023}
}

@inproceedings{DBLP:conf/emnlp/MinKLLYKIZH23,
  author       = {Sewon Min and
                  Kalpesh Krishna and
                  Xinxi Lyu and
                  Mike Lewis and
                  Wen{-}tau Yih and
                  Pang Wei Koh and
                  Mohit Iyyer and
                  Luke Zettlemoyer and
                  Hannaneh Hajishirzi},
  title        = {FActScore: Fine-grained Atomic Evaluation of Factual Precision in
                  Long Form Text Generation},
  booktitle    = {{EMNLP}},
  pages        = {12076--12100},
  publisher    = {Association for Computational Linguistics},
  year         = {2023}
}

\appendix

\section{Experimental Details}

\subsection{Data Construction Details}

To construct the supervised fine-tuning dataset, we set the temperature to 0 and adopt GPT-3.5 Turbo. We concatenate question $q$ and its related subgraph $G'$ using a prompt template to form the input $x$. ChatGPT generates candidate knowledge representation $k$ based on 3 examples as demonstrations and input $x$ under ICL paradigm. Given that, the objective of $k$ is to augment the performance of the QA model, we evaluate the quality of $k$ by examining the QA model's results. We utilize $k$ as contextual knowledge for the QA model to generate the answer $a$. If the answer $a$ encompasses all the answer entities, knowledge $k$ is considered helpful for answering the question, and $(x,k)$ is used as an input-output pair for supervised training.

For the construction of the preference dataset, we sample knowledge representations from the knowledge rewriter after supervised fine-tuning. We set the temperature to 1 to foster greater diversity. During preference annotation, given that the knowledge rewriter aims to generate contextual knowledge beneficial for QA, we first assess the quality of knowledge representation based on the number of answer entities they contain. This na\"{i}ve strategy is robust and cost-effective, avoiding additional API calls for evaluations using ChatGPT, thus saving time and reducing costs. We label the candidate with the highest number of answer entities as the preferred knowledge representation $k^+$ and the one with the fewest as the dispreferred $k^-$. If the number of answer entities is the same, we select the two, $k_1$ and $k_2$, with the greatest semantic difference by using all-MiniLM-L6-v2\footnote{https://huggingface.co/sentence-transformers/all-MiniLM-L6-v2} as the encoder. This selection process ensures a significant semantic gap between the two chosen representations, facilitating more rapid model convergence during training. Subsequently, $k_1$ and $k_2$ serve as contextual knowledge to prompt the QA model, yielding answers $a_1$ and $a_2$. We annotate the preferred knowledge representation $k^+$ and the dispreferred knowledge representation $k^-$ by evaluating the quality of $a_1$ and $a_2$ using ChatGPT. Finally, ChatGPT is used to paraphrase the preferred knowledge representation $k^+$ into an enhanced version $k^{++}$, forming a preference pair $k^{++}$ and $k^-$ for direct preference optimization (DPO).

\subsection{Datasets}
\textbf{GrailQA}\footnote{This dataset is distributed under the CC BY-SA 4.0 license and our utilization complies with the terms specified in the license.} \cite{DBLP:conf/www/GuKVSLY021} is a challenging, large-scale multi-hop KGQA benchmark. It is an English dataset that utilizes Freebase \cite{DBLP:conf/aaai/BollackerCT07,DBLP:conf/sigmod/BollackerEPST08} as KG. It spans 86 domains, such as Sports, Location, and Computer Video Games, and comprises 64,331 questions (44,337 train, 6,763 dev, 13,231 test). This dataset features a large number of entities and relations, complex logical forms, and noise in entity mentions within the questions. The training and dev sets provide annotated SPARQL queries and answer entities, while the test set comprises only the questions. For evaluation convenience, the dev set is used for testing. 

\noindent\textbf{GraphQuestions}\footnote{This dataset is licensed under the Creative Commons Attribution 4.0 and our usage aligns with the intended purposes outlined in this license.} \cite{DBLP:conf/emnlp/SuSSSGYY16} is a characteristic-rich dataset for factoid question answering based on Freebase across 70 domains, like People, Astronomy, and Medicine. This English dataset focuses on the following question characteristics: structure complexity, function, commonness, paraphrasing, and answer cardinality. It comprises 5,166 questions (2,771 train, 2,395 test), with nearly half requiring multi-hop reasoning. For each question, the dataset provides corresponding SPARQL query and answer entities.

\subsection{Large Language Models}
\label{Large Language Models}
\textbf{Llama-2}\footnote{The license of Llama-2 is available at https://ai.meta.com/resources/models-and-libraries/llama-downloads/.} \cite{touvron2023llama2}, an updated version of Llama-1 \cite{DBLP:journals/corr/abs-2302-13971}, is developed using a training corpus comprising 2 trillion tokens and features a context length twice that of Llama-1. To better accomplish the knowledge rewriting task, we select Llama-2-7B-Chat\footnote{https://huggingface.co/meta-llama/Llama-2-7b-chat-hf}.

\noindent\textbf{Llama-3}\footnote{The license of Llama-3 is available at https://llama.meta.com/llama3/license.} \cite{llama3modelcard} is the latest model in the Llama series. It is renowned for its mastery of language nuances, contextual comprehension, and proficiency in executing complex tasks such as translation and generating dialogues. We choose Llama-3-8B-Instruct\footnote{https://huggingface.co/meta-llama/Meta-Llama-3-8B-Instruct} for knowledge rewriting.

\noindent\textbf{Mistral}\footnote{It is under the Apache 2 License.} \cite{jiang2023mistral} is an open-source LLM developed by Mistral AI. We select the latest instruction-tuned version, Mistral-7B-Instruct-v0.3\footnote{https://huggingface.co/mistralai/Mistral-7B-Instruct-v0.3}, as our QA model.

\noindent\textbf{ChatGPT}\footnote{The terms of use for ChatGPT are available at https://openai.com/policies/terms-of-use/.}, developed by OpenAI, is a milestone in the era of LLMs. Its robust capabilities in natural language understanding and generation facilitate superior performance across various tasks. We leverage GPT-3.5-turbo via the API\footnote{https://api.openai.com/} for knowledge rewriting and question answering.

All the LLMs above are general-domain models. Regarding language support,  Llama-2, Llama-3, and Mistral only support English, while ChatGPT is multilingual. In this study, the use of these LLMs complies with their respective licenses or terms.

\begin{table}[h]
\small
\scalebox{1}{%
\begin{tcolorbox}
    \textbf{KG-to-Text Prompt}\\
    \\
    \textcolor{orange}{\textbf{[Instruction]}}\\
    Your task is to transform a knowledge graph to a sentence or multiple sentences. The knowledge graph is: \{triples\}. The sentence is:\\
    \rule{\linewidth}{0.2mm}
    \textbf{Summary Prompt}\\
    \\
    \textcolor{orange}{\textbf{[Instruction]}}\\
    Your task is to summarize the relevant knowledge that is helpful to answer the question from the following triples.\\
    \textbf{Triples:} \{triples\}\\
    \textbf{Question:} \{question\}\\
    \textbf{Knowledge:}\\
    \rule{\linewidth}{0.2mm}
    \textbf{CoTKR Prompt}\\
    \\
    \textcolor{orange}{\textbf{[Instruction]}}\\
    Your task is to summarize the relevant information that is helpful to answer the question from the following triples. Please think step by step and iteratively generate the reasoning chain and the corresponding knowledge.\\
    \textbf{Triples:} \{triples\}\\
    \textbf{Question:} \{question\}
\end{tcolorbox}
}
\caption{Prompts for Knowledge Rewriting Methods.}
\label{Table 4}
\end{table}

\begin{table}[h]
\small
\scalebox{1}{%
\begin{tcolorbox}
    \textbf{Prompt for Question Answering with Triple/KG-to-Text/Summary Knowledge}\\
    \\
    \textcolor{orange}{\textbf{[Instruction]}}\\
    Your task is to answer the question based on the knowledge that might be relevant. Try to use the original words from the given knowledge to answer the question. But if it is not useful, just ignore it and generate your own guess.\\
    \textbf{Knowledge:} \{knowledge\}\\
    \textbf{Question:} \{question\}\\
    \textbf{Answer:}\\
    \rule{\linewidth}{0.2mm}
    \textbf{Prompt for Question Answering with
CoTKR/CoTKR+PA Knowledge}\\
    \\
    \textcolor{orange}{\textbf{[Instruction]}}\\
    Your task is to answer the question based on the reasoning chain that might be relevant. Try to use the original words from the given knowledge to answer the question. But if it is not useful, just ignore it and generate your own guess.\\
    \textbf{Knowledge:} \{knowledge\}\\
    \textbf{Question:} \{question\}\\
    \textbf{Answer:}\\
    \rule{\linewidth}{0.2mm}
    \textbf{Prompt for Question Answering without Context}\\
    \\
    \textbf{Question:} \{question\}\\
    \textbf{Answer:}\\
\end{tcolorbox}
}
\caption{Prompts for Question Answering.}
\label{Table 5}
\end{table}

\begin{table}[h]
\small
\scalebox{1}{%
\begin{tcolorbox}
    \textbf{Preference Annotation Prompt}\\
    \\
    \textcolor{orange}{\textbf{[Instruction]}}\\
    Your task is to evaluate the quality of two responses to the question based on predefined criteria. Avoid any position biases and ensure that the order in which the responses were presented does not influence your decision. Do not allow the length of the responses to influence your evaluation. Be as objective as possible.\\
    \textcolor{darkred}{\textbf{[Criteria]}}\\
    For this evaluation, you should primarily consider the following criteria:\\
    \textbf{Accuracy:} The response should contain as many answer entities as possible, and use the original words of the answer entities.\\
    \textbf{Relevance:} The response should be to the point of the question.\\
    \textbf{Question:} \{question\}\\
    \textbf{Answer:} \{answer\}\\
    \textbf{Response A:} \{response A\}\\
    \textbf{Response B:} \{response B\}\\
    \textcolor{darkgreen}{\textbf{[Evaluation Rule]}}\\
    Begin your evaluation by comparing the two responses and provide a short explanation. Then output only the single character: "A" if Response A is better, "B" if Response B is better, and "C" for a tie. At the end, repeat just the letter again by itself on a new line.\\
\end{tcolorbox}
}
\caption{Preference Annotation Prompt.}
\label{Table 6}
\end{table}

\subsection{Retrieval Methods Details}
\label{retrieve}

\textbf{2-Hop} subgraph is a na\"{\i}ve question-related context. Most KBQA studies under RAG paradigms consider triples within the N-hop subgraph of the head entity as contextual knowledge \cite{baek-etal-2023-knowledge,sen-etal-2023-knowledge,DBLP:journals/corr/abs-2401-00426,DBLP:journals/corr/abs-2403-02966}. To retrieve the 2-hop subgraph around the head entity, we execute SPARQL queries on Freebase. Given the large size of the 2-hop subgraph, we use all-MiniLM-L6-v2\footnote{https://huggingface.co/sentence-transformers/all-MiniLM-L6-v2} to encode all 1-hop and 2-hop relations of the head entity and the question, excluding meaningless relations, such as ``common.topic.webpage''. Then, we select semantically similar relation paths based on cosine similarity. Finally, we sample the corresponding triples from KG based on these relation paths. In this experiment, we select the top 30 triples as our retrieval results. However, the small size of the 2-hop subgraphs for some entities may result in fewer than 30 triples being retrieved.

\noindent\textbf{BM25} \cite{DBLP:journals/ftir/RobertsonZ09} is a retrieval method based on TF-IDF scores of sparse word matching between input questions and passages. This method is commonly used for integrating multimodal data sources or text-based QA \cite{DBLP:journals/corr/abs-2308-13259,DBLP:conf/iclr/YuZNZL0HWWX23}. We follow the processing method in DecAF \cite{DBLP:conf/iclr/YuZNZL0HWWX23}, simply linearizing the 1-hop subgraph of the topic entity as the article. We use the BM25 implemented by Pyserini \cite{DBLP:conf/sigir/LinMLYPN21} and collect the triples corresponding to the candidate articles as the retrieval result. Specifically, we initially retrieve 10 candidate articles, each containing up to 10 triples. Subsequently, we remove any redundant triples or those containing meaningless relations. Given the limited context length of LLMs, we select the top 30 triples as the context information for question answering. After filtering, the number of triples in the candidate articles may be less than 30, thus resulting in the retrieval subgraphs for some questions containing fewer than 30 triples.

\noindent\textbf{Ground Truth Subgraph (GS)} refers to a subgraph consisting of the triples necessary for answering a question. In this experiment, we modify the SPARQL queries provided in the datasets and execute them on Freebase to obtain the ground truth subgraphs. We use this subgraph to represent the results of an ideal retriever, aiming to explore the performance upper bound of different knowledge rewriting strategies for the QA model.
\begin{table}[h]
\small
\scalebox{1}{%
\begin{tcolorbox}
    \textbf{Data Augmentation Prompt}\\
    \\
    \textcolor{orange}{\textbf{[Instruction]}}\\ 
    You are a knowledge graph summarizer for Question Answering. I will give you "Question", "Triple", "Answer" and "Knowledge". Your task is to paraphrase the original "Knowledge" into a more helpful representation format for Question Answering. The "Paraphrased Knowledge" should contain the original words of all the answer entities.\\
    \textbf{Question:} \{question\}\\
    \textbf{Triple:} \{triples\}\\
    \textbf{Knowledge:} \{knowledge\}\\
    \textbf{Paraphrased Knowledge:}\\
\end{tcolorbox}
}
\caption{Data Augmentation Prompt.}
\label{Table 7}
\end{table}

\subsection{Implementation Details}

We utilize LoRA \cite{DBLP:conf/iclr/HuSWALWWC22} to achieve parameter-efficient fine-tuning. For supervised fine-tuning and DPO, the batch size, learning rate, lora rank, lora alpha, and lora dropout are set to 128, 1e-4, 64, 128, and 0.05, respectively. In supervised fine-tuning, we train for 10 epochs and save the best model based on validation set results. In DPO, we observe that more training steps may lead to decreased model performance. Consequently, we train for 1 to 2 epochs on GraphQuestions using approximately 2,000 training samples, and for 5 to 20 steps on GrailQA using 640 to 2560 training samples. During inference, the temperature is set to 0 for ChatGPT and 0.01 for open-source LLMs, ensuring relatively stable output. All the parameter settings mentioned above are the optimal results after multiple trials.

\begin{table*}[h]
\centering
\small
\scalebox{0.8}{
\begin{tabular}{lp{7cm}p{7cm}}
    \toprule
    \multicolumn{3}{l}{\textbf{Question}: what is the unit of area that the measurement system that have an electric field strength units of \textcolor{darkgreen}{volt per metre} have?}\\
    \multicolumn{3}{l}{\textbf{Answer}: \textcolor{orange}{Square kilometer}, \textcolor{orange}{Square meter}}\\  
    \midrule
    \textbf{KR Methods}& \multicolumn{1}{c}{\textbf{KR Result}} & \multicolumn{1}{c}{\textbf{QA Response}}\\
    \midrule
    \multirow{5}{*}{\textbf{Triple}}& (International System of Units, electric\_field\_strength\_units, \textcolor{darkgreen}{Volt per metre}) (International System of Units, area\_units, \textcolor{orange}{Square meter}) (International System of Units, area\_units, \textcolor{orange}{Square kilometer}) (\textcolor{darkgreen}{Volt per metre}, strength\_in\_volts\_per\_meter, 1.0) ... & \textcolor{orange}{Square meter}.\\
    \cdashline{1-3}
    \multirow{7}{*}{\textbf{KG-to-Text}} & The International System of Units includes various units for different physical quantities such as length, area, potential, substance, conductance, volume, resistance, current, force, and pressure. For example, the unit for electric field strength in \textcolor{darkgreen}{volts per meter} is 1.0. Additionally, it covers a wide range of units including \textcolor{orange}{square kilometers}, \textcolor{orange}{square meters} ...&The unit of area in the International System of Units that has an electric field strength unit of \textcolor{darkgreen}{volt per meter} is \textcolor{orange}{square meters} (m²).\\
    \cdashline{1-3}
    \multirow{7}{*}{\textbf{Summary}} & The International System of Units includes various measurement systems for different physical quantities such as length, area, volume, conductance, resistance, current, force, and pressure. The measurement system with an electric field strength unit of \textcolor{darkgreen}{Volt per meter} does not directly relate to a specific area unit within the International System of Units. & There is no direct relationship between the measurement system with an electric field strength unit of \textcolor{darkgreen}{Volt per meter} and any specific area unit within the International System of Units.\\
    \cdashline{1-3}
    \multirow{10}{*}{\textbf{CoTKR}} & Reason: I need to know the unit of area associated with the measurement system that has an electric field strength unit of \textcolor{darkgreen}{Volt per meter}. \newline Summarize: The International System of Units includes various units for different physical quantities such as length, area, volume, etc.\newline Reason: I need to find the specific unit of area in the International System of Units that corresponds to the electric field strength unit of \textcolor{darkgreen}{Volt per meter}.\newline Summarize: The International System of Units includes units like \textcolor{orange}{Square kilometer}, \textcolor{orange}{Square meter}, etc. for measuring area.&The unit of area in the measurement system that has an electric field strength unit of \textcolor{darkgreen}{Volt per meter} is \textcolor{orange}{Square meter} (m²).\\
    \cdashline{1-3}
    \multirow{6}{*}{\textbf{CoTKR+PA}}&Reason: I need to know which measurement system includes the unit for electric field strength in \textcolor{darkgreen}{Volt per meter}.\newline Knowledge: The International System of Units contains the unit for electric field strength in \textcolor{darkgreen}{Volt per meter}.\newline Reason: I need to identify the unit of area within the International System of Units.\newline Knowledge: The International System of Units includes the units \textcolor{orange}{Square kilometer} and \textcolor{orange}{Square meter} as area units.& The unit of area within the International System of Units that the measurement system having an electric field strength unit of \textcolor{darkgreen}{volt per metre} has is either \textcolor{orange}{Square kilometer} or \textcolor{orange}{Square meter}.\\
    \bottomrule
\end{tabular}
}
\caption{An example of knowledge rewriting results for different methods. We use Llama-3 and Mistral as the Knowledge Rewriting (KR) backbone and the QA model, respectively.}
\label{Table 8}
\end{table*}

Our implementation utilizes PyTorch\footnote{https://pytorch.org/}, Transformers\footnote{https://huggingface.co/docs/transformers/en/index}, DeepSpeed\footnote{https://www.deepspeed.ai/}, Datasets\footnote{https://huggingface.co/docs/datasets/en/index}, PEFT\footnote{https://huggingface.co/docs/peft/en/index}, and TRL\footnote{https://huggingface.co/docs/trl/en/index}. We use Datasets for data preprocessing. Both training and inference are based on PyTorch and Transformers. Supervised fine-tuning and DPO are implemented using PEFT, TRL, and DeepSpeed. Experiments are conducted on 4 NVIDIA A100-SXM4-40GB GPUs, with each training or inference session completed within one day. Due to the high computational costs of LLMs, we conduct each experiment once and then report the results.

All the prompts involved in this experiment are as follows. Table \ref{Table 4} shows the prompts for different knowledge rewriting methods. Table \ref{Table 5} shows the prompts for question answering. Table \ref{Table 6} shows the preference annotation prompt.
\label{PA}

\subsection{Prompt for Data Augmentation}
\label{data augmentation}
Table \ref{Table 7} shows the data augmentation prompt.

\section{Case Study}
\label{case}

In this section, we compare different knowledge rewriting strategies through an example. As illustrated in Table \ref{Table 8}, the knowledge generated by both \textsf{Triple} and \textsf{KG-to-Text} contains excessive redundant information. This redundancy complicates the process for the QA model, making it challenging to extract relevant knowledge. \textsf{Summary} struggles to extract useful information when faced with an abundance of triples. In contrast, \textsf{CoTKR} and \textsf{CoTKR+PA} summarize the most pertinent knowledge in the rewriting step, thereby enabling the QA model to provide a concise and accurate answer. Furthermore, after preference alignment, our knowledge rewriter is capable of generating more natural
reasoning steps, significantly enhancing its applicability to KGQA.

\section{Additional Experimental Results}

\subsection{Experiments on ComplexWebQuestions}
To evaluate the robustness of CoTKR, we conduct our experiments on ComplexWebQuestions\cite{DBLP:conf/naacl/TalmorB18}. We utilize ChatGPT as the knowledge rewriter, Mistral as the question-answering model, and 2-Hop as the retrieval method. The experimental results, presented in Table \ref{Table 9}, demonstrate the effectiveness of CoTKR.

\begin{table}[h]
\centering
\small
\scalebox{1}{
\begin{tabular}{lccc}
    \toprule
     \textbf{Methods} & \textbf{Acc} & \textbf{Recall} & \textbf{EM}\\
    \midrule
     \textbf{No Knowledge}& 36.82 &31.30& 	27.62\\
     \textbf{Triple}& 39.29 &33.94 &30.86\\
     \textbf{KG-to-Text}& 35.96 &31.53 &28.96\\
     \textbf{Summary}& 38.53 &33.87 &30.89\\
     \textbf{CoTKR}& \textbf{40.70} &\textbf{35.74} &\textbf{32.72}\\
    \bottomrule
\end{tabular}
}
\caption{Experiments on ComplexWebQuestions use ChatGPT as the knowledge rewriter, Mistral as the QA model, and 2-Hop for retrieval.}
\label{Table 9}
\end{table}

\subsection{Knowledge Rewriter with GPT-4}
To assess the applicability of CoTKR to GPT-4, we further conduct the experiments with GPT-4 as the knowledge rewriter, Mistral as the question-answering model, and 2 hop as the retrieval method on 1,000 test questions from GrailQA. The detailed results are presented in Table \ref{Table 10}. The findings show that CoTKR outperforms other approaches, with CoTKR utilizing GPT-4 achieving the highest performance. This suggests that employing a more advanced LLM backbone, such as GPT-4, leads to superior outcomes.

\begin{table}[h]
\centering
\small
\scalebox{1}{
\begin{tabular}{lccc}
    \toprule
     \textbf{Methods} & \textbf{Acc} & \textbf{Recall} & \textbf{EM}\\
    \midrule
     \textbf{No Knowledge}& 29.10 &21.87&18.80\\
     \textbf{Triple}& 54.30 &47.15 &42.40\\
     \textbf{KG-to-Text}& 53.20 &45.42 &40.60\\
     \textbf{Summary}& 56.00 &48.63 &43.60\\
     \textbf{CoTKR}& \textbf{57.50} &\textbf{52.16} &\textbf{48.20}\\
    \bottomrule
\end{tabular}
}
\caption{Experiments on 1,000 GrailQA test questions use GPT-4 for rewriting, Mistral for QA, and 2-Hop as the retrieval method.}
\label{Table 10}
\vspace{-0.2cm}
\end{table}

\subsection{Time Analysis}
We conduct experiments to analyze the average time cost of the knowledge rewriting methods discussed in this paper. We adopt Llama-3 as the knowledge rewriter, Mistral as the question-answering model, and 2-Hop as the retrieval method. The experiments are conducted on GraphQuestions, utilizing one A100-SXM4-40GB GPU. The average runtime for each question by different methods is shown in Table \ref{Table 11} (unit: seconds). The average runtime of each question for all methods is within an acceptable range (i.e., less than 1.5 seconds). Although our method is the most time-consuming, it exhibits a clear advantage in performance.

\begin{table}[!h]
\centering
\small
\resizebox{\columnwidth}{!}{
\begin{tabular}{lccccc}
    \toprule
     \textbf{Process} & \textbf{No Knowledge} & \textbf{Triple} & \textbf{KG-to-Text} &\textbf{Summary} & \textbf{CoTKR}\\
    \midrule
     \textbf{Rewrite}& 0& 0 &1.0828 & 0.4615& 0.9835\\
     \textbf{Answer}& 0.3787& 0.5998 &0.2701 & 0.2502& 0.4414\\
     \textbf{Rewrite+Answer}& 0.3787& 0.5998 &1.3529 & 0.7117& 1.4249\\
    \bottomrule
\end{tabular}
}
\caption{Time analysis on GraphQuestions (seconds) using Llama-3 as the knowledge rewriter, Mistral for question answering, and 2-Hop for retrieval.}
\label{Table 11}
\end{table}

\subsection{GPT-4-score as Evaluation Metrics}
Given the powerful natural language understanding and generation capabilities of closed-source large models, many existing works employ ChatGPT as an evaluator to provide high-quality evaluation results \cite{DBLP:conf/emnlp/SottanaLZY23,DBLP:conf/emnlp/LiuIXWXZ23,DBLP:conf/emnlp/MinKLLYKIZH23}. In our approach, we use GPT-4 as the evaluator to assess whether all answer entities are present in the responses. We refer to this evaluation metric as \textbf{GPT-4-score}. Compared to EM, this metric is more flexible, as LLMs are capable of recognizing synonyms of answer entities. We use it to evaluate the first 300 questions from GrailQA using Llama-3 as knowledge rewriter, ChatGPT as question-answering model, and 2 hop as retrieval method. We provide the prompt for \textbf{GPT-4-score} in Table \ref{Table 12}.

\begin{table}[h]
\small
\scalebox{1}{%
\begin{tcolorbox}
    \textbf{GPT-4-score Prompt}\\
    \\
    \textcolor{orange}{\textbf{[Instruction]}}\\
    Your task is to evaluate the quality of the response to the question. You should consider whether all the answer entities appear in the response.\\
    \textbf{Question:} \{question\}\\
    \textbf{Answer:} \{answer\}\\
    \textbf{Response:} \{response\}\\
    Begin your evaluation by comparing the response and the answer and provide a short explanation. Then output only the single number: "1" if all the answer entities appear in the response, and "0" if not. At the end, repeat just the number again by itself on a new line.
\end{tcolorbox}
}
\caption{GPT-4-score Prompt.}
\label{Table 12}
\end{table}

The experimental results are shown in Table \ref{Table 13}. The results indicate that our implemented GPT-4-score is effective and consistent with the outcomes reflected by other evaluation metrics. Furthermore, this also demonstrates that CoTKR possesses significant advantages compared to other knowledge rewriting methods.

\begin{table}[!h]
\centering
\small
\scalebox{1.2}{
\begin{tabular}{lc}
    \toprule
     \textbf{Methods} & \textbf{GPT-4-score}\\
    \midrule
     \textbf{No Knowledge}& 0.2767\\
     \textbf{Triple}& 0.4700\\
     \textbf{KG-to-Text}& 0.4749\\
     \textbf{Summary}& 0.4800\\
     \textbf{CoTKR}& 0.5167\\
     \textbf{CoTKR+PA}& \textbf{0.5567}\\
    \bottomrule
\end{tabular}
}
\caption{GPT-4-score on GraphQuestions. We use Llama-3 as knowledge rewriter, ChatGPT as question-answering model, and 2 hop as retrieval method.}
\label{Table 13}
\end{table}

\subsection{Qualitative Analysis on Data Augmentation}
To clearly demonstrate the importance of data augmentation, we perform a qualitative analysis. The results reveal that data augmentation enhances performance in three key areas: reducing redundant reasoning steps, supplementing critical information, and generating more concise summaries. These improvements are exemplified by three representative cases, as shown in Table \ref{Table 14}.
\begin{table*}[h]
\centering
\small
\scalebox{0.8}{
\begin{tabular}{lp{4cm}p{5cm}p{5cm}}
    \toprule
    \textbf{Effect} & \textbf{Question}& \boldsymbol{$k^+$} & \boldsymbol{$k^{++}$}\\ 
    \midrule
    \multirow{11}{3cm}{\textbf{Avoiding Redundant Reasoning Step}} & Katy börner is the curator for what exhibition?& \underline{Reason}: I need to know which exhibition Katy Börner curated.\newline \underline{Summarize}: Katy Börner curated the exhibition "Places \& Spaces".\newline \underline{Reason}: I need to know the subjects of the exhibition "Places \& Spaces" curated by Katy Börner.\newline \underline{Summarize}: The subjects of the exhibition "Places \& Spaces" curated by Katy Börner include Cartography, Information visualization, Information design, and Scientometrics. & \underline{Reason}: I need to know which exhibition Katy Börner curated. \newline \underline{Summarize}: Katy Börner curated the exhibition "Places \& Spaces".\\
    \cline{1-4}
    \multirow{6}{3cm}{\textbf{Key Points Supplementation}} & Person of interest deals with how many tv subjects? & \underline{Reason}: I need to know how many TV subjects are associated with "Person of Interest". \newline \underline{Summarize}: "Person of Interest" is associated with multiple TV subjects such as Espionage, Mass surveillance, Hacker, and Vigilante. & \underline{Reason}: I need to know how many TV subjects are associated with "Person of Interest". \newline \underline{Summarize}: "Person of Interest" deals with a total of 5 TV subjects including Espionage, Mass surveillance, Hacker, Surveillance, and Vigilante.\\
    \cline{1-4}
    \multirow{4}{3cm}{\textbf{More Concise Summarization}} & Who are the owners of gree? & \underline{Reason}: I need to know who owns GREE, Inc.\newline \underline{Summarize}: GREE, Inc. is owned by GREE, Inc. itself, according to the triple (GREE, internet.website.owner, GREE, Inc.).& \underline{Reason}: I need to know who owns GREE, Inc. \newline \underline{Summarize}: GREE, Inc. is owned by GREE, Inc. itself.\\
    \bottomrule
\end{tabular}
}
\caption{Representative cases for improvements through data augmentation.}
\label{Table 14}
\end{table*}

\subsection{Detailed Experimental Results}
This section presents all the experimental results of this study. As shown in Table \ref{Table 15}, Table \ref{Table 16}, and Table \ref{Table 17}, CoTKR/CoTKR+PA achieves the best performance in most scenarios. This indicates that CoTKR is effective for both open-source LLMs after training and closed-source LLMs using ICL. Besides, the results also reveal the robustness of CoTKR, demonstrating its applicability across KGQA systems with various retrieval methods and QA models.

\begin{table*}[h]
\centering
\small
\scalebox{0.8}{
\begin{tabular}{llccccccc}
    \toprule
     \multirow{2}{*}{\textbf{KR LLMs}} & \multirow{2}{*}{\textbf{Methods}} & \multicolumn{3}{c}{\textbf{GrailQA}} & &\multicolumn{3}{c}{\textbf{GraphQuestions}}\\
    \cline{3-5}
    \cline{7-9}
     & & \textbf{Acc} & \textbf{Recall} & \textbf{EM}& & \textbf{Acc} & \textbf{Recall} & \textbf{EM}\\
    \midrule
    \multicolumn{9}{c}{\textit{ChatGPT as QA model}}\\
    \multirow{2}{*}{\textbf{None}} & No Knowledge & 28.91& 22.81& 20.14& &35.87 &25.76 &22.09 \\
    & Triple &57.76 & 49.67&44.73 & &55.03 &46.65 &41.63\\
    \cdashline{1-9}
    \multirow{4}{*}{\textbf{Llama-2}} & KG-to-Text &54.75 & 47.35&42.44 & &49.73 &40.00 &33.74 \\
    & Summary &58.14 &51.38 &46.38 & &\underline{52.94} &44.70 &38.41\\
     & CoTKR &\underline{58.64} &\underline{52.33} &\underline{47.88} & &51.36 &\underline{45.20} &\underline{39.96}\\
     & CoTKR+PA & \textbf{59.25} &\textbf{53.52} &\textbf{49.64} & &\textbf{56.78} &\textbf{47.99} &\textbf{42.46}\\
    \cdashline{1-9}
     \multirow{4}{*}{\textbf{Llama-3}} & KG-to-Text &55.76 &48.41 &43.90 & &52.40 &45.06 &39.83\\
     & Summary &57.55 &51.06 &46.80 & &\underline{54.95} &46.86 &40.75\\
     & CoTKR &\underline{58.33} &\underline{52.55} &\underline{48.65} & &53.19 &\underline{47.23} &\underline{43.17}\\
     & CoTKR+PA &\textbf{61.51} & \textbf{56.08}& \textbf{52.67} & &\textbf{56.37} &\textbf{49.31} & \textbf{45.26}\\
    \cdashline{1-9}
     \multirow{3}{*}{\textbf{ChatGPT}} & KG-to-Text &56.32 &49.05 &44.73 & &53.53 &45.59 &41.17\\
     & Summary &\underline{58.54} &\underline{51.81} &\underline{47.29} & &\textbf{55.62} &\textbf{48.93} &\textbf{44.97}\\
     & CoTKR &\textbf{59.87} &\textbf{53.19} &\textbf{49.02} & &\underline{54.28} &\underline{48.18} &\underline{44.68}\\
    \midrule
    \multicolumn{9}{c}{\textit{Mistral as QA model}}\\
     \multirow{2}{*}{\textbf{None}} & No Knowledge &29.44 &23.13 &20.30 & &38.20 &26.92 &22.13 \\
     & Triple &54.47 &47.78 &43.25 & & 51.32& 45.97&41.67\\
    \cdashline{1-9}
     \multirow{4}{*}{\textbf{Llama-2}} & KG-to-Text &49.49 &42.91 &38.41 & &44.59 & 37.98&32.82 \\
     & Summary &54.10 &47.79 &43.15 & &49.85 &42.33 &36.45\\
     & CoTKR &\underline{56.75} &\underline{51.10} & \underline{46.71} & &\underline{50.19} &\underline{43.73} &\underline{38.54}\\
     & CoTKR+PA &\textbf{58.15} & \textbf{52.98}&\textbf{49.13} & &\textbf{55.07} &\textbf{47.02} &\textbf{41.71}\\
    \cdashline{1-9}
     \multirow{4}{*}{\textbf{Llama-3}} & KG-to-Text &50.64 &44.32 &40.13 & &49.06 & 43.04&38.25\\
     & Summary &53.84 &47.71 &43.49 & &52.03 & 44.30&38.50\\
     & CoTKR &\underline{56.47} &\underline{51.33} &\underline{47.36} & & \underline{52.65}& \underline{46.48}&\underline{42.21}\\
     & CoTKR+PA &\textbf{59.31} &\textbf{54.13} &\textbf{50.24}& &\textbf{54.82} & \textbf{47.76}&\textbf{43.09}\\
    \cdashline{1-9}
     \multirow{3}{*}{\textbf{ChatGPT}} & KG-to-Text &51.04 &44.87 &40.97 & &49.14 &43.04 &38.83\\
     & Summary &\underline{54.44} & \underline{48.16} &\underline{43.97} & &\underline{52.28} &\underline{47.10} &\underline{43.30}\\
     & CoTKR &\textbf{57.28} &\textbf{51.14} &\textbf{47.09} & & \textbf{52.82}&\textbf{47.13} &\textbf{43.55}\\
    \bottomrule
\end{tabular}
}
\caption{The overall results of CoTKR and the baselines on GrailQA and GraphQuestions using 2-Hop as retrieval method. For each combination of the Knowledge Rewriter (KR) LLM and the QA model, the best and second-best results are highlighted in bold and underlined, respectively.}
\label{Table 15}
\end{table*}

\begin{table*}[h]
\small
\centering
\scalebox{0.8}{
\begin{tabular}{llccccccc}
    \toprule
     \multirow{2}{*}{\textbf{KR LLMs}} & \multirow{2}{*}{\textbf{Methods}} & \multicolumn{3}{c}{\textbf{GrailQA}} & &\multicolumn{3}{c}{\textbf{GraphQuestions}}\\
    \cline{3-5}
    \cline{7-9}
     & & \textbf{Acc} & \textbf{Recall} & \textbf{EM}& & \textbf{Acc} & \textbf{Recall} & \textbf{EM}\\
    \midrule
    \multicolumn{9}{c}{\textit{ChatGPT as QA model}}\\
    \multirow{2}{*}{\textbf{None}} & No Knowledge & 28.91& 22.81& 20.14& &35.87 &25.76 &22.09 \\
    & Triple &58.42 & 49.06&43.87 & &48.43 &40.44 &36.83\\
    \cdashline{1-9}
    \multirow{4}{*}{\textbf{Llama-2}} & KG-to-Text &52.80 & 43.93&39.01 & &41.67 &32.09 &27.68 \\
    & Summary &57.62 &49.13 &43.80 & &\underline{46.47} &36.84 &31.44\\
     & CoTKR &\textbf{58.32} &\underline{50.32} &\underline{45.57} & &45.34 &\underline{37.21} &\underline{32.73}\\
     & CoTKR+PA & \textbf{58.32} &\textbf{50.77} &\textbf{46.41} & &\textbf{49.52} &\textbf{40.06} &\textbf{35.16}\\
    \cdashline{1-9}
     \multirow{4}{*}{\textbf{Llama-3}} & KG-to-Text &55.14 &46.64 &41.96 & &44.68 &34.51 &30.19\\
     & Summary &\underline{59.19} &50.83 &46.19 & &44.97 &36.97 &32.44\\
     & CoTKR &58.89 &\underline{51.13} &\underline{46.93} & &\underline{46.18} &\underline{38.59} &\underline{35.03}\\
     & CoTKR+PA &\textbf{59.69} & \textbf{52.34}& \textbf{48.14} & &\textbf{50.44} &\textbf{42.03} & \textbf{37.58}\\
    \cdashline{1-9}
     \multirow{3}{*}{\textbf{ChatGPT}} & KG-to-Text &55.61 &47.00 &42.38 & &\textbf{46.14} &36.64 &31.98\\
     & Summary &\underline{58.76} &\underline{50.47} &\underline{45.66} & &\textbf{46.14} &\textbf{39.20} &\textbf{35.66}\\
     & CoTKR &\textbf{59.66} &\textbf{51.06} &\textbf{46.40} & &45.18 &\underline{38.03} &\underline{34.36}\\
    \midrule
    \multicolumn{9}{c}{\textit{Mistral as QA model}}\\
     \multirow{2}{*}{\textbf{None}} & No Knowledge &29.44 &23.13 &20.30 & &38.20 &26.92 &22.13 \\
     & Triple &55.58 &47.10 &42.07 & & 43.34& 36.98&33.86\\
    \cdashline{1-9}
     \multirow{4}{*}{\textbf{Llama-2}} & KG-to-Text &47.33 &39.90&35.64 & &35.24 & 28.35&25.05 \\
     & Summary &52.94 &44.94 &40.12 & &43.47 &34.56 &29.52\\
     & CoTKR &\underline{55.69} &\underline{48.37}& \underline{43.77} & &\underline{45.39} &\underline{37.07} &\underline{32.73}\\
     & CoTKR+PA &\textbf{57.13} & \textbf{50.17}&\textbf{45.94} & &\textbf{49.02} &\textbf{39.46} &\textbf{34.53}\\
    \cdashline{1-9}
     \multirow{4}{*}{\textbf{Llama-3}} & KG-to-Text &49.83 &42.40 &38.07& &39.21 & 31.86&28.48\\
     & Summary &54.33 &46.73 &42.35 & &41.63 & 34.41&30.15\\
     & CoTKR &\underline{56.76} &\underline{49.75} &\underline{45.57} & & \underline{44.38}& \underline{37.14}&\underline{33.57}\\
     & CoTKR+PA &\textbf{58.27} &\textbf{51.00} &\textbf{46.72}& &\textbf{48.14} & \textbf{39.34}&\textbf{34.91}\\
    \cdashline{1-9}
     \multirow{3}{*}{\textbf{ChatGPT}} & KG-to-Text &50.35 &42.69 &38.40 & &39.46 &33.36 &29.52\\
     & Summary &\underline{54.98} & \underline{47.15} &\underline{42.63} & &\underline{41.42}&\underline{35.60}&\underline{31.86}\\
     & CoTKR &\textbf{56.84} &\textbf{48.87} &\textbf{44.17} & & \textbf{43.42}&\textbf{36.23} &\textbf{32.57}\\
    \bottomrule
\end{tabular}
}
\caption{The overall results of CoTKR and the baselines on GrailQA and GraphQuestions using BM25 as retrieval method. For each combination of the Knowledge Rewriter (KR) LLM and the QA model, the best and second-best results are highlighted in bold and underlined, respectively.}
\label{Table 16}
\end{table*}

\begin{table*}[h]
\centering
\small
\scalebox{0.8}{
\begin{tabular}{llccccccc}
    \toprule
     \multirow{2}{*}{\textbf{KR LLMs}} & \multirow{2}{*}{\textbf{Methods}} & \multicolumn{3}{c}{\textbf{GrailQA}} & &\multicolumn{3}{c}{\textbf{GraphQuestions}}\\
    \cline{3-5}
    \cline{7-9}
     & & \textbf{Acc} & \textbf{Recall} & \textbf{EM}& & \textbf{Acc} & \textbf{Recall} & \textbf{EM}\\
    \midrule
    \multicolumn{9}{c}{\textit{ChatGPT as QA model}}\\
    \multirow{2}{*}{\textbf{None}} & No Knowledge & 28.91& 22.81& 20.14& &35.87 &25.76 &22.09 \\
    & Triple &77.41 & 67.14&61.47 & &83.17 &70.89 &62.76\\
    \cdashline{1-9}
    \multirow{4}{*}{\textbf{Llama-2}} & KG-to-Text &80.05 & 70.20&65.06 & &76.24 &64.47 &57.58 \\
    & Summary &85.14 &75.90 &70.03 & &\underline{80.88} &69.31 &60.00\\
     & CoTKR &\underline{87.18} &\underline{78.46} &\underline{73.56} & &80.67 &\underline{70.73}&\underline{63.72}\\
     & CoTKR+PA & \textbf{87.79} &\textbf{79.86} &\textbf{75.50} & &\textbf{83.67}&\textbf{75.02}&\textbf{68.85}\\
    \cdashline{1-9}
     \multirow{4}{*}{\textbf{Llama-3}} & KG-to-Text &82.17 &73.10 &68.28 & &78.66 &68.32 &59.92\\
     & Summary &85.44 &77.01 &72.07 & &83.55 &72.47 &64.47\\
     & CoTKR &\underline{88.27} &\underline{80.35} &\underline{75.93} & &\underline{85.51} &\underline{75.89}&\underline{69.14}\\
     & CoTKR+PA &\textbf{91.48} & \textbf{84.02}& \textbf{79.93} & &\textbf{87.85} &\textbf{79.66} & \textbf{73.95}\\
    \cdashline{1-9}
     \multirow{3}{*}{\textbf{ChatGPT}} & KG-to-Text &80.39 &71.19&66.46 & &83.05 &73.60 &67.22\\
     & Summary &\underline{86.53} &\underline{77.88} &\underline{72.87} & &\underline{88.39} &\underline{79.26} &\underline{72.99}\\
     & CoTKR &\textbf{87.77} &\textbf{78.97} &\textbf{74.38} & &\textbf{89.81} &\textbf{80.09}&\textbf{73.49}\\
    \midrule
    \multicolumn{9}{c}{\textit{Mistral as QA model}}\\
     \multirow{2}{*}{\textbf{None}} & No Knowledge &29.44 &23.13 &20.30 & &38.20 &26.92 &22.13 \\
     & Triple &74.12 &65.83 &60.79 & & 83.26& 72.17&64.22\\
    \cdashline{1-9}
     \multirow{4}{*}{\textbf{Llama-2}} & KG-to-Text &72.47 &64.18&59.53 & &71.19 & 59.62&51.98 \\
     & Summary &78.87 &70.43 &64.71 & &\underline{78.71} &67.45 &58.29\\
     & CoTKR &\underline{84.27} &\underline{76.57}& \underline{71.60} & &78.66&\underline{68.75} &\underline{61.80}\\
     & CoTKR+PA &\textbf{85.81} & \textbf{79.26}&\textbf{74.52} & &\textbf{81.38} &\textbf{72.88} &\textbf{67.06}\\
    \cdashline{1-9}
     \multirow{4}{*}{\textbf{Llama-3}} & KG-to-Text &74.54 &66.58 &62.38 & &76.37 &66.56&58.75\\
     & Summary &79.64 & 71.65 &66.89& &80.84&70.07&62.21\\
     & CoTKR &\underline{84.73} &\underline{77.86} &\underline{73.31} & & \underline{84.18}&\underline{75.04}&\underline{68.64}\\
     & CoTKR+PA &\textbf{87.80} &\textbf{81.24} &\textbf{76.65}& &\textbf{84.63} & \textbf{77.38}&\textbf{73.24}\\
    \cdashline{1-9}
     \multirow{3}{*}{\textbf{ChatGPT}} & KG-to-Text &72.53 &65.06&60.92 & &81.71 &72.35&65.43\\
     & Summary &\underline{81.34} & \underline{72.53} &\underline{67.53}& &\underline{86.47}&\underline{77.15}&\underline{70.56}\\
     & CoTKR &\textbf{84.77} &\textbf{76.72} &\textbf{72.11} & & \textbf{88.60}&\textbf{78.95}&\textbf{72.36}\\
    \bottomrule
\end{tabular}
}
\caption{The overall results of CoTKR and the baselines on GrailQA and GraphQuestions using GS as retrieval method. For each combination of the Knowledge Rewriter (KR) LLM and the QA model, the best and second-best results are highlighted in bold and underlined, respectively.}
\label{Table 17}
\end{table*}

\section{AI Assistants in Research or Writing}
In this research, we primarily utilize ChatGPT for the construction of training data and as the foundational model for knowledge rewriting and QA. For academic writing, ChatGPT is used to correct grammatical errors.

\end{document}